\documentclass[conference]{IEEEtran}
\IEEEoverridecommandlockouts
\usepackage{cite}
\usepackage{amsmath}

\newcommand{\ts}[1]   {_{\text{#1}}}
\newcommand{\tb}[1]   {_{\textbf{#1}}}

\usepackage{amssymb,amsfonts}

\usepackage{algorithmic}
\usepackage{graphicx}
\usepackage{textcomp}
\usepackage{xcolor}
\usepackage{multirow}
\usepackage{gensymb} 
\usepackage{tabularx}
\usepackage{ulem}
\usepackage{euscript}

\def\BibTeX{{\rm B\kern-.05em{\sc i\kern-.025em b}\kern-.08em
    T\kern-.1667em\lower.7ex\hbox{E}\kern-.125emX}}
\begin{document}

\title{\textbf{UMArm}: \textbf{U}ntethered, \textbf{M}odular, Portable, Soft Pneumatic \textbf{Arm}\\}

% Author list version that saves space
\author{Runze Zuo, %<-this % stops a space
    Dong Heon Han, %
    Richard Li, %
    Saima Jamal, %
    Daniel Bruder%
\thanks{The authors are with the Mechanical Engineering Department at the University of Michigan, Ann Arbor, MI 48109, USA \{\tt\small zuorunze, dongheon, richieli, saimaj, bruderd\}@umich.edu}%
}
%~~~~~~~~~~~~~~~~end preamble~~~~~~~~~~~~~~~~~%

% TITLE
\maketitle

\begin{abstract}
% Robotic arms play a crucial role in modern industries, but their adaptability to unstructured environments remains a challenge. While rigid robotic arms excel in precision, speed, and payload capacity, their reliance on highly structured spatial information limits their usability in dynamic settings. Soft robotic arms, particularly pneumatically actuated designs, offer improved adaptability and safety for human-robot interactions. However, current pneumatic soft arms face significant drawbacks, including the need for external compressors, complex tubing systems, and uncontrolled compliant degrees of freedom, which hinder their precision and portability. In this work, we present a novel pneumatically driven rigid-soft hybrid robotic arm that uses valve-embedded McKibben actuators (VEMAs) to drive a rigid spine. The embedded valve feature of VEMAs allowed us to eliminate individual pressure lines for each actuator, significantly reducing system weight and complexity. We were able to power the robot arm using a lightweight carbon fiber air tank, making it completely untethered, demonstrating wear-ability and assist-ability. The modular structure allows easy reconfiguration; in a separate demo, we showed that the system can be reconfigured into a snake-like robot for complex terrain navigation. Our results highlight the potential of this hybrid approach to enhance portability, dexterity, and adaptability for real-world applications in unstructured environments.
%
%
% passive voice version
Robotic arms in modern industries lack adaptability to unstructured environments. Soft pneumatic arms offer better adaptability and human-robot safety but suffer from limited degrees of freedom, precision, payload capacity, and reliance on bulky external regulators. This work presents "UMArm," a novel pneumatic rigid-soft hybrid arm that addresses these limitations through densely integrated, self-regulated McKibben actuators on a lightweight rigid spine. The modified actuators incorporate valves and controllers internally, eliminating individual pressure lines and external regulators, significantly reducing system weight and complexity. UMArm achieves full untethered operation, high payload capacity, precision, and tunable compliance. Portability is demonstrated via wearable assistive arm experiments, and versatility through reconfiguration into an inchworm robot. Results show that high-degree-of-freedom, external-regulator-free pneumatic arm systems like UMArm have great potential for real-world unstructured environments.
\end{abstract}

%% ABSTRACT

%%%%%% For inline comments %%%%%%
\newcommand{\dan}[1]{{\normalsize{\textbf{({\color{blue}Dan:\ }#1)}}}}

\newcommand{\needcite}[1]{{\normalsize{\textbf{({\color{blue}cite\ }#1)}}}}
\newcommand{\runze}[1]{{\normalsize{\color{blue}\textbf{(#1)}}}}
\newcommand{\don}[1]{{\normalsize{\color{green}\textbf{(#1)}}}}

%%%%%% For first round revisions (show in red) %%%%%%
\newcommand{\revcomment}[2]{\textcolor{red}{#2}$^{\##1}$} % Include reference commands
\newcommand{\tempcomment}[2]{{#2}$^{\##1}$}    % for comments that should only show on the marked up version

\renewcommand{\revcomment}[2]{#2}   % removes coloring of edited sections, i.e. makes doc look normal (comment this line out for highlighted edits)
\renewcommand{\tempcomment}[1]{}    % removes the comment completely

%%%%%% Other commands %%%%%%

%% BODY
\section{Introduction}

\begin{figure}
    \centering
    \includegraphics[width=\linewidth]{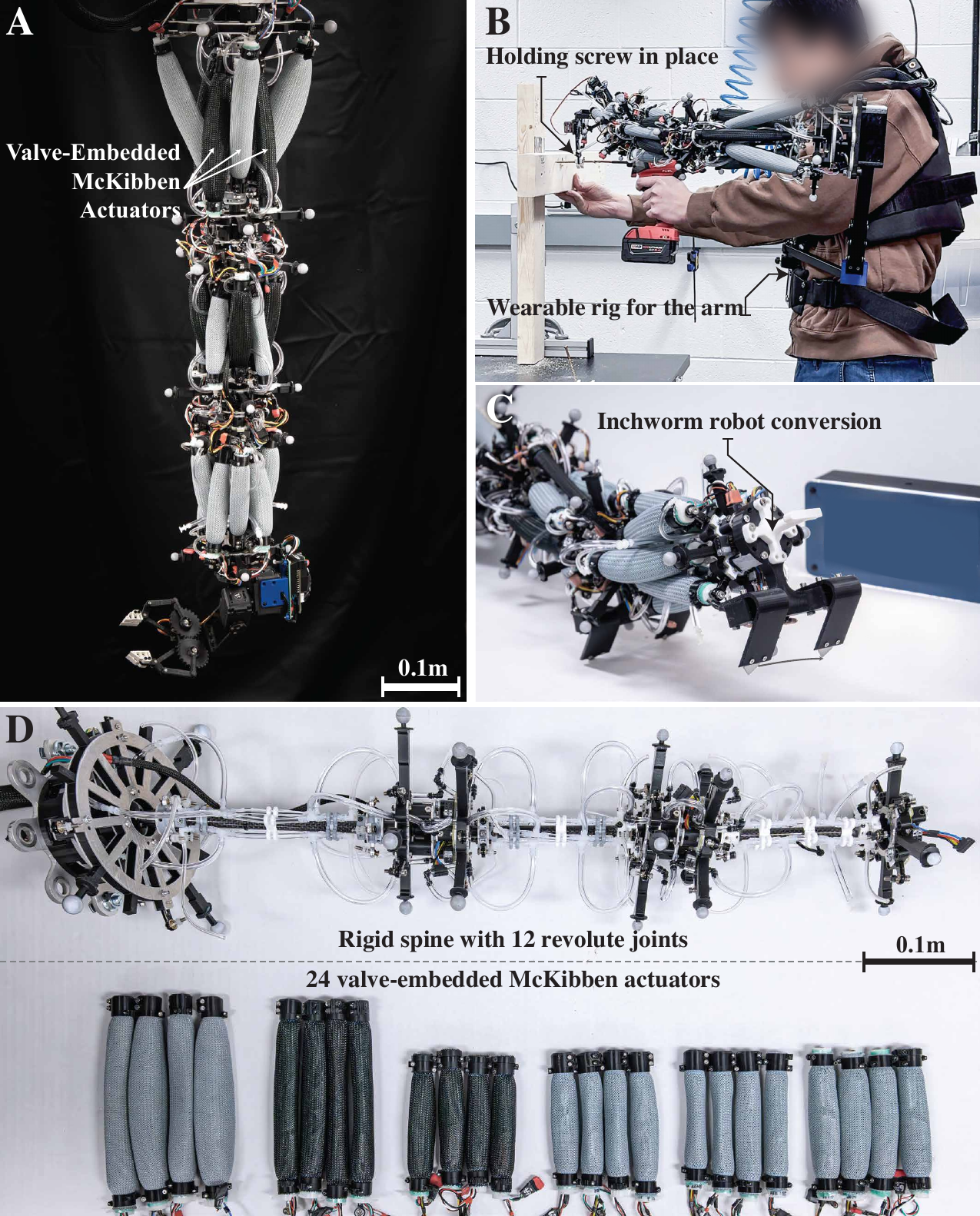}
    \caption{\textbf{A:} Photo of the proposed rigid-soft pneumatically driven arm. It has 24 independently controlled, self-regulating actuators, and weighs 1.15kg. \textbf{B:} The arm is wearable and can assist with carpentry tasks. In the figure, it holds a screw in place while a human worker drills it in with an impact driver. \textbf{C:} With slight modification in between the arm segments, the robot arm can perform snake-like locomotion. It operates untethered with pressure supplied from a lightweight carbon fiber tank. \textbf{D:} Rigid 12-DoF spine and valve-embedded McKibben actuators}
    \vspace{-0.5cm}
    \label{fig:overview}
\end{figure}

Robotic arms are essential components in modern industries. Rigid robotic arms can operate with great precision, speed, and payload capacity, making them ideal for highly structured environments like factories \cite{almurib2012review,brogaardh2007present,kemp2007challenges}. However, their application in unstructured real-world environments remains limited due to their lack of compliance. In unstructured environments, compliance is crucial for safe interaction and effective handling of unexpected disturbances. Compliance can be achieved either by structure or by control. Rigid robots can simulate compliance through control, but it requires the coordinated control of all actuators, and will be limited by controller bandwidth and sensor latency. Without full-body compliance, rigid arms have to operate on highly precise and low-latency spatial information—a requirement that is difficult to satisfy in unstructured environments \cite{zheng2020robust, wang2022control}. Consequently, rigid arms can cause damage to target objects, nearby humans, or the robot itself \cite{yang2019collaborative,zhang2020modular,le2020challenges}.

Unlike their rigid counterparts, soft robotic arms have structural compliance. They are capable of mitigating external disturbances and compensating for misalignment through elastic deformation of their structure, making them more suitable for operation in unstructured environments \cite{trimmer2013soft,rus2015design,li2024disturbance}. Their inherent compliance also ensures safety in physical human-robot interaction, enabling use cases such as daily assistance and elder care \cite{ansari2017towards,firth2022anthropomorphic}.

A variety of soft actuation strategies have been explored to enable compliance and safety in unstructured environments. HASEL actuators achieve high strain and power output via electrohydraulic zipping, enabling fast, modular actuation; however, their high-voltage requirements and structural fragility pose challenges for unstructured deployment \cite{mitchell2019easy}. Shape memory alloy (SMA)-driven soft actuators offer compact and lightweight actuation with high power-to-weight ratios and low voltage requirements but suffer from slow thermal response, nonlinear behavior, and sensitivity to ambient temperature—factors that hinder precise control \cite{yang2018design}. Hydraulic soft actuators provide high force output\cite{feng2021design}, yet their practicality is limited by their bulky fluidic components and slow responses. Tendon-driven designs have also gained attention, offering enhanced workspaces and tunable compliance through mechanisms like trimmed helicoids \cite{guan2023trimmed}, but these systems remain vulnerable to friction, slack, and lack robustness under dynamic conditions \cite{guan2023trimmed,yang2018design}. 

\revcomment{2.2}{
\begin{table*}[t]
\centering
\caption{Comparison of UMArm with similar-scale soft robotic arm designs from the literature 
% Unlike prior works, UMArm is fully untethered while simultaneously achieving portability, precision, payload capacity, and dexterity.
}
\label{tab:comparison}
\renewcommand{\arraystretch}{1.2}
\begin{tabularx}{\textwidth}{l c c X X}
\hline
\textbf{Work} & \textbf{Untethered} & \textbf{DoFs} & \textbf{Weight / Force Output} & \textbf{Position Control Accuracy} \\
\hline
Poly-Limb (Nguyen et al.)\cite{Nguyen2019Soft} & Tethered & $9\ts{continuum}$ & 1.6 kg(Arm only) / 0.3~kg payload & No closed loop position control \\
Lightweight PAM Arm (Ohta et al.) \cite{ohta2018design} & Tethered & $5\ts{Soft}+2\ts{Rigid}$ & 1.2 kg(Arm only) / 0.5~kg payload & Joint level control only \\
Stiffening Arm (Bruder et al.) \cite{bruder2023increasing} & Tethered & $9\ts{continuum}$ & 1 kg(Arm only) / 0.5~kg payload & No closed loop position control \\
Koopman Arm (Bruder et al.) \cite{bruder2021koopman} & Tethered & $6\ts{Variable Stiffness}$ & 1 kg(Arm only) / 0.275~kg payload & Koopman-MPC, 30mm~RMSE \\
\textbf{UMArm (this work)} & \textbf{Untethered} & \textbf{$12\ts{Variable Stiffness}$} & \textbf{1.15 kg (with pressure regulators) / 53.3N blocked force} & \textbf{Closed loop inverse kinematics control, 0.85~mm RMSE} \\
\hline
\end{tabularx}
\end{table*}
}

% \subsection*{pneumatically driven soft arms are most promising for deployment to unstructured real world environment}

Among various soft actuation strategies, pneumatic actuators are promising due to their smooth, continuous deformation, high force-to-weight ratio, and compliance in every direction. Unlike cable-driven mechanisms, pneumatic actuators do not require complex cable routing, which simplifies design and thus enhances reliability. Compared to dielectric elastomers, HASEL actuators, and shape memory alloys, pneumatic actuators offer a more practical path toward building human-arm-scale systems that avoids the need for high-voltage operation, complex material packaging, or thermally limited cycle rates. 
These properties, combined with their resilience to impact, ease of fabrication, and cost-effectiveness, make pneumatic actuators particularly well-suited for building robot arms for deploying into real-world, unstructured environments \cite{whitesides2018soft,xavier2022soft,garcia2024design,galloway2016soft,trivedi2008soft,seleem2023recent}.

% \subsection*{shortcomings}

Although pneumatically actuated soft arms have desirable characteristics, certain drawbacks limit their performance. Firstly, although many pneumatic soft arms are lightweight, they require bulky external pressure regulators \cite{sokolov2023design}. The cumbersome external hardware limits their portability. Secondly, high dexterity pneumatic soft arms need many independently actuated degrees of freedom (DoF), but the centralized, external regulating scheme makes it difficult to increase independent DoFs. This is because adding independently controlled pneumatic actuators requires extra external regulators and more tubing running from the actuators to the external regulators. Large amounts of tubing will introduce extra weight, thereby affecting the payload capacity of the arm. At the same time, long thin tubes increase fluidic resistance, causing pressure to propagate more slowly from the regulator to the actuator, affecting pressure control performance \cite{zuo2024embedded}.
Thirdly, structurally soft pneumatic arms might suffer from uncontrolled compliant degrees of freedom (DoF). Under heavy load, they might buckle at unwanted positions, affecting precision and payload capacity \cite{bruder2020data}. Moreover, robots might oscillate at the uncontrolled compliant joints when given aggressive control inputs \cite{li2021position}. 
% In summary, future research should focus on improving portability, increasing the number of independently actuated DoFs, and precision and payload capacity to build pneumatic soft arms deployable to unstructured environments.
% In summary, current pneumatically driven arms lack the portability, dexterity, precision, and payload capacity to operate effectively.

\revcomment{4.3}{
Researchers have explored various pneumatic soft arm architectures to address common limitations. Continuum soft pneumatic arms \cite{hongwei2022kinematic, alessi2023ablation, zhu2024design, jiang2020design, garcia2024design, ian2006octarm}, which offer infinite degrees of freedom (DoFs) and desirable traits such as flexibility and low cost. They often use silicone rubber or origami structures. However, their uncontrolled DoFs reduce precision and payload capacity. }

\revcomment{4.3}{
To overcome this, rigid–soft hybrid arms have been developed, combining rigid bones and joints with soft pneumatic actuators. This reduces uncontrolled compliance by limiting it to the joints. For instance, Ohta et al. created a rigid-bone arm actuated by McKibben actuators \cite{ohta2018design}. While being lightweight, only five of its seven DoFs are actuated by McKibbens, and the low contraction ratio of these actuators introduces a tradeoff between joint force and range of motion. As a result, McKibben-based arms with limited DoFs struggle to achieve both large payload capacity and workspace size. Adding more independently controlled actuators could help, but increases design complexity due to additional tubing and regulators.  
}

\revcomment{4.3}{
Several hybrid designs have also been proposed to improve force transmission and scalability. Bruder et al. addressed payload limitations with antagonistic stiffening in a McKibben-based soft arm \cite{bruder2023increasing}, improving force transmission. However, each segment uses only one actuator for stiffening, requiring all DoFs in the segment to stiffen together, which limits motion during stiffening. Origami-based hybrid designs are also promising. Liu et al. used water bomb origami in a rigid–soft arm \cite{liu2023design}, achieving high payload capacity, though the locking mechanism only engages when fully extended. Oh et al. used origami bellows as actuators \cite{oh2024hybrid}, producing strong force output and motion range, but with limited DoFs and scaling challenges. Zhou et al. developed a compact segment with many independently controlled rubber bellows \cite{zhou2024pneumatic}, achieving excellent dexterity and precision. However, the scale of the robot is relatively small, the possibility of scaling up is unknown, and, like all existing designs, it relies on bulky, centralized regulators that reduce portability and hinder use in unstructured environments.  
}

%%% Dan's suggested rewrite of contributions part of intro
To address the key limitations of existing pneumatically driven soft arms -- namely, limited precision, payload capacity, and reliance on bulky external regulators -- we propose a new design approach that integrates pressure regulation directly into the actuators. Building on a recent work \cite{zuo2024embedded}, where the authors introduced the valve-embedded McKibben actuator (VEMA), we present \textbf{UMArm}, a pneumatically driven rigid-soft hybrid robotic arm designed for high performance in unstructured environments.

% VEMAs incorporate onboard valves for internal pressure regulation, allowing multiple actuators to be connected to a shared pneumatic-power-communication bus. This architecture eliminates the need for individual pneumatic lines and electrical connections for each actuator, enabling a clean, compact, and lightweight system. This makes it possible to scale up the number of actuators without compromising portability or increasing complexity.

% UMArm leverages these capabilities to create 
UMArm is a highly capable robotic arm with 24 independently controlled actuators, untethered operation capability, and the ability to \revcomment{0}{directionally generate a 53.3N force} while weighing only 1.15\,kg. Its design combines embedded actuator pressure regulation, a dense arrangement of independent actuators, tunable stiffness, and rigid structural support within a portable form factor. Specifically, this work introduces three key hardware innovations:
\begin{itemize}
    \item Integrated pneumatic-power-communication bus architecture, enabling dozens of independently controlled pneumatic actuators to be operated in parallel with shared tubing and wiring.
    \item High-density actuator arrangement that accommodates eight VEMAs per arm segment, allowing for multidirectional antagonistic stiffening at every joint.
    \item Lightweight articulated rigid spine structure for efficient force transmission despite a large range of motion.
\end{itemize}
UMArm’s design directly addresses the core challenges faced by pneumatically driven soft arms. First, it enhances portability and removes dependence on bulky external regulators by employing VEMAs, which integrate local pressure control within each actuator. This distributed control eliminates the need for centralized pneumatic hardware and individual pneumatic lines for each actuator. Second, the pneumatic-power-communication bus architecture, combined with the dense arrangement of VEMAs, allows for a high number of independently actuated degrees of freedom (DoFs). This enables precise, fine-grained control and supports directional compliance. Third, the rigid spine is engineered to withstand the significant forces produced by antagonistic actuator pairs during stiffening, resulting in improved force transmission and a high strength-to-weight ratio. Finally, by incorporating antagonistic actuation at each joint, UMArm provides tunable compliance. This enables the arm to selectively modulate stiffness in different directions, improving adaptability and precision when interacting with complex or unpredictable environments.

\revcomment{2.2, 4.3}{
To better situate our contribution, Table~\ref{tab:comparison} summarizes representative soft robotic arms from the literature, focusing on systems of comparable scale to UMArm. Early continuum and McKibben-based designs demonstrated impressive compliance, dexterity, or payload capacity, but lacked closed-loop precision and remained tethered to external regulators. More recent works introduced stiffening mechanisms or advanced control strategies (e.g., Koopman-based MPC), which improved load handling and tracking but still relied on bulky hardware and achieved only centimeter-level endpoint accuracy. In contrast, UMArm is fully untethered, portable, and precise (0.85\,mm RMSE), with high force output and 12 actively controlled DoFs with tunable stiffness. This combination of features distinguishes UMArm from prior soft robotic arms and enables high-performance operation in unstructured environments.
}

To demonstrate UMArm's effectiveness, we experimentally validated its ability to perform tasks that have traditionally been challenging for pneumatically driven soft arms. \revcomment{0}{First, we conducted a series of experiments to evaluate UMArm's precision, force-generation capability, and untethered operation. Precision was assessed through a position-control step response experiment and a trajectory-tracking experiment. The step response experiment achieved a positional control RMSE of 0.85\,mm, while the trajectory-tracking experiment showed a near-identical match between the target and actual trajectories. UMArm's force-generation ability was demonstrated with a tested output up to 53.3\,N.} Additionally, we tested its untethered operation capability by powering it with a portable compressed air tank, which successfully sustained the UMArm for 12 minutes at full capacity. Second, through the use of antagonistic stiffening, UMArm achieves high-precision manipulation. It can stack three small objects with great consistency. Moreover, the antagonistic stiffness is independently controlled for each joint, granting UMArm the ability to perform directional compliance control. We have shown that the UMArm can structurally stiffen in one direction while remaining soft in another, enabling more adaptive interactions. Third, we demonstrate that UMArm can be worn and used on worksites, highlighting both its portability and its capability of using its compliant structure to compensate for misalignment and disturbances. Finally, to showcase its versatility, we added legs between arm segments, transforming UMArm into an untethered inchworm robot, further broadening its potential application space.

\section{Design}
\begin{figure}
    \centering
    \includegraphics[width=\linewidth]{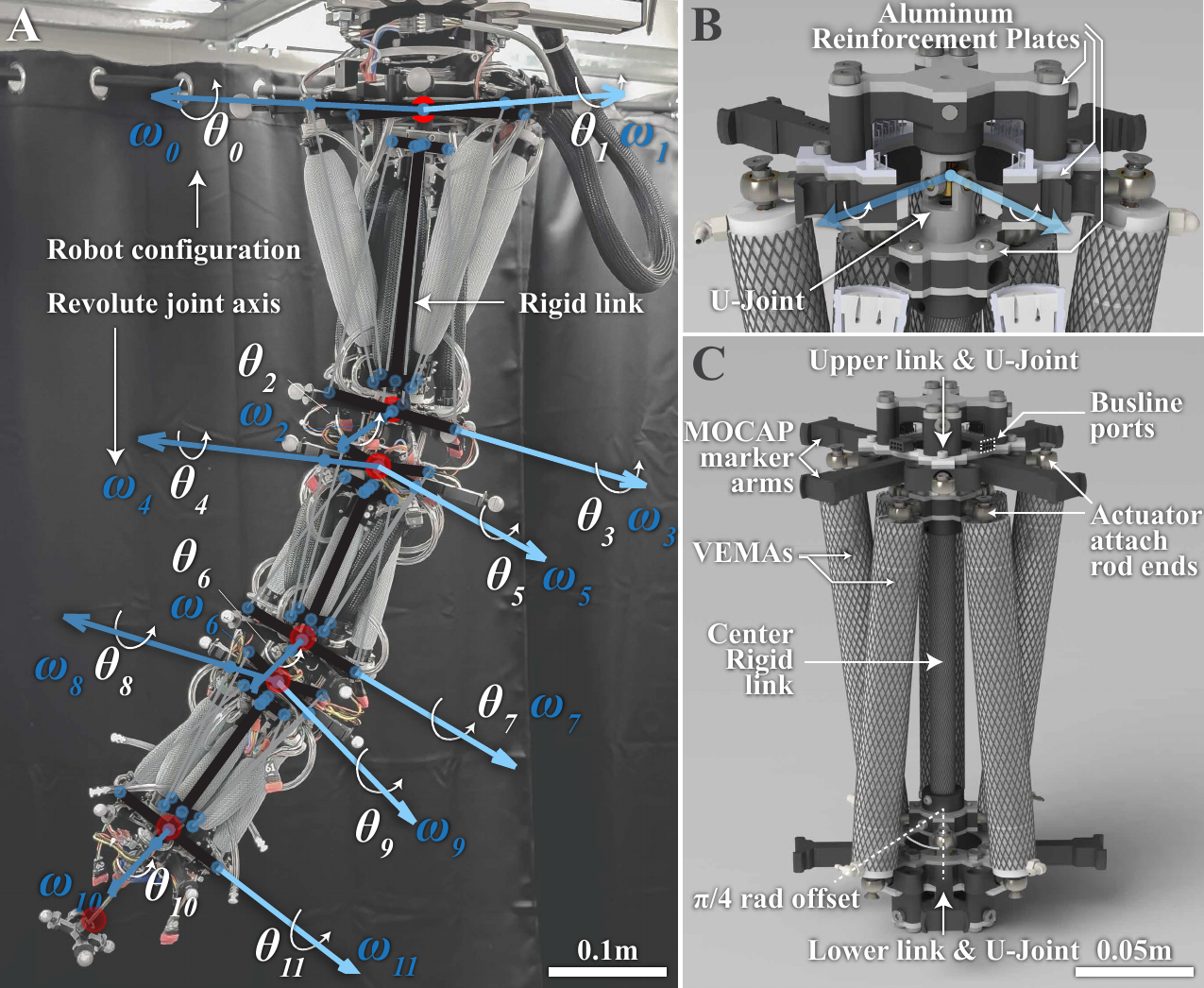}
    \caption{\textbf{A:} Photo of the robot arm overlaying the annotation of the revolute axes. There are 12 revolute joints in total for all three segments, controlled by 24 actuators. \textbf{B:} Close-up rendered picture of a joint. A U-joint connects two adjacent rigid links. Each U-joint is modeled as 2 revolute joints. \textbf{C:} Rendered Picture showing the structure of one segment of UMArm. There are two groups (of 4) of actuators. Each group controls the two revolute joints within one U-joint. Two groups are stacked reversely with $\pi/4$ rad offset angle.}
    \label{fig:structure}
\end{figure}

In designing the UMArm, we aim to explore a method for scaling up the number of independently controlled actuators within a compact system and to determine what performance gains can be achieved through this approach. UMArm is a hybrid rigid-soft arm. UMArm has a rigid spine driven by soft actuators. The rigidity of a spine significantly improves precision and force output, and the soft actuators maintain the structural compliance of the arm. As illustrated in figure \ref{fig:overview}D, the UMArm consists of three \revcomment{1.8}{structurally similar spine segments, with decreasing length from base to tip}. Each segment has three rigid-body links — an upper plate, a central rod, and a lower plate — connected in series by two universal joints (U-joints). We chose to use U-joints because they are lightweight and have no backlash. The U-joints also restrict z-axis rotation, which is helpful since we do not have actuators to control that degree of freedom. Each U-joint (Fig.~\ref{fig:structure}B), can be simplified as 2 revolute joints with perpendicular axes. As a result, the UMArm can be seen as a robot with 12 revolute joints, assuming that the links between the two joints are rigid.

To let the system accommodate a greater number of independently controlled actuators, we employ the valve-embedded McKibben actuator (VEMA) that features self-pressure regulating \cite{zuo2024embedded}. All VEMAs are connected to the shared constant pressure source, which can be a single pressurized tube with a larger diameter that runs through the arm. Therefore, we eliminated the need for individual long tubing that connects each actuator to its external regulator. Compared to using the thin, long individual tubing for every actuator, the use of shared, larger-diameter tubing will decrease the whole system's weight and complexity and improve pressure control performance by lowering the pneumatic friction. Also, after establishing the shared pressure line, adding more actuators to the arm would need very little additional hardware, making it possible to scale up the number of actuators in our arm. Under this pressure bus line architecture, we have installed 24 individually controlled VEMAs to the arm. Having such a large number of independently controlled actuators gives us the ability to simultaneously control the joint's angle and structural stiffness for all 12 joints. Structural stiffness control is achieved through antagonistically placing the actuators for each joint. In the following sections of this manuscript, we will demonstrate that being able to control the stiffness of every joint can improve payload capacity, arm precision, and safety while working around humans.

\iffalse
Bruder et al. demonstrated the benefits of antagonistic stiffening in a McKibben actuated arm \cite{bruder2023increasing}. Still, due to insufficient independent control channels, the antagonistic module can only uniformly assert pressure to all bending directions, affecting the robot's posture. Building on this concept, the UMArm utilizes two VEMAs per joint to form an antagonistic pair, thereby enabling effective antagonistic stiffening while maintaining the original joint angles.
\fi

Limited joint range of motion remains a significant challenge for McKibben-based antagonistic stiffening arms. McKibben actuators can only contract by approximately 34\% \cite{bruder2021chain}. When used to drive a revolute joint on an arm, one can increase the range of motion of the joint by decreasing the length of the lever arm, but at the cost of decreasing torque output. UMArm can address this limitation by incorporating more joints within the same arm length, because although each joint would have a small range of motion (but larger force), the net range of motion across all the joints can remain large. Figure \ref{fig:structure}C shows the actuator placement. We integrated 8 VEMAs into one arm segment. Four VEMAs control the two joint angles between the upper link and the center link. In the resting state, these actuators are angled toward the center. This ensures better force output at a larger joint angle, where the contracted actuator will have better alignment with the force lever arm, and the elongated actuator will generate larger forces at a given pressure, in accordance with the McKibben model in \cite{bruder2018force}. The gap between the four upper VEMAs can accommodate another group of four VEMAs that drive the joint angles between the center rod and lower plate. As shown in Fig.\ref{fig:structure}C, the lower group is reversed and rotated by $\pi/4$ rad to prevent interference while operating. This configuration of VEMA placement efficiently utilizes the space, thus increasing the density of joints, which, as a result, effectively improves the overall range of motion. The UMArm is built with 3 of these segments. The base segment is designed to withstand more load. It has a longer force lever arm for the VEMAs and uses a variant of VEMA with a larger diameter, which will output larger force compared to the smaller diameter VEMAs at the same length and pressure.

The bulky pressure regulating hardware is distributed into every VEMAs in the arm, therefore, even with a total of 24 independently controlled VEMAs, the UMArm does not need any external regulators. The input to the arm only consists of a single 276kPa(40 psi) pneumatic line and a 4-wire power and communication bus line. This simplicity allows untethered operation. In this work, we use a paintball compressed air tank and a step-down regulator to provide the 276kPa(40 psi) pneumatic pressure, a small Li-ion battery pack to supply 12VDC-2A for the arm's electronics, and a Bluetooth-RS485 communication module to receive commands from the host PC wirelessly.

\section{Kinematic Model and Control}

\subsection{Forward Kinematics of the UMArm}
The arm is modeled as a 12-revolute joint robot.  In this work, we employ the standard product of exponential method \cite{murray2017mathematical}. For each U-joint, We define the two revolute joints as illustrated in Fig. \ref{fig:structure}B. The rotation axes of each of the 12 revolute joints can be expressed as $\overrightarrow{\omega} = [\omega_{0}, ..., \omega_{11}]^\intercal, \omega_{i}\in \mathbb{R}^3$, and we choose $q_i$ as the center of the U-joint, where two revolute axes intersect: $\overrightarrow{q} = [q_{0}, ..., q_{11}]^\intercal, q_i \in \mathbb{R}^3$. The rotation angle variable for each revolute joint is $\overrightarrow{\theta} = [\theta_{0}, ..., \theta_{11}]^\intercal, \theta_i \in \mathbb{R}$. The 3 segments of the arm have identical structure, therefore, we can first compute the forward kinematics for each segment consists of 4 joints using $\overrightarrow{\omega}$, $\overrightarrow{q}$, and $\overrightarrow{\theta}$:

\begin{equation}
\begin{cases}
    \xi_{i} &= [-\omega_i \times q, \omega_i]^\intercal \in \mathbb{R}^6 \\
    g_{st}^{seg1}(\theta_{0-3}) &= e^{\xi_{0}\theta_{0}} e^{\xi_{1}\theta_{1}} e^{\xi_{2}\theta_{2}} e^{\xi_{3}\theta_{3}} g_{st}^{seg1}(0)
\end{cases}
\label{twistexp}
\end{equation}

$g_{st}^{seg1}(0) \in SE(3)$ represents the relative transformation from the base-frame to the tool-frame at robot initial posture, where all $\theta=0$. $\xi_{i}$ is the twist of a joint, which contains a three-dimensional angular velocity vector and a three-dimensional linear velocity vector. By using the $\xi_{i}$ and the joint angle $\theta$, we can construct the twist exponential $e^{\xi_{i}\theta_{i}}$. Using the twist exponential for each joint, and the formulation in Equation \ref{twistexp}, we can acquire $g_{st}^{seg1}(\theta_{0-3}) : \mathbb{R}^4\rightarrow SE(3)$, which is the forward kinematics of one arm segment. It maps from a set of joint angles to the tool frame expressed in the segment base coordinate frame.

We can assemble a full-robot forward kinematics by using the forward kinematics for each segment. The robot's initial posture ($\overrightarrow\theta = 0$) is shown in Fig. \ref{fig:overview}A: the first segment is mounted on the ceiling and all the links align with the spatial Z-axis. For each segment, the base-frame is the upper link plate in Fig. \ref{fig:structure}C, and the tool-frame is the lower link plate. The origin of both frames coincides with the center of the U-joints. We assume that the base-frame of the robot is also the base-frame of the first segment. $H_{s_i}^{s_{i+1}} \in SE(3)$ is the link joint offset, i.e., the homogeneous transformation between the current segment tool-frame and the next segment base-frame.

Since the robot consists of three segments, we assume that the tool-frame of the entire robot is the tool-frame of the $3^{rd}$ segment. We can then express the entire robot forward kinematics $g_{st}^{robot}(\theta) : \mathbb{R}^{12} \rightarrow SE(3)$ as:

\revcomment{4.5}{
\begin{equation}
    g_{st}^{robot}(\theta) =g_{st}^{seg1}(\theta_{0-3})H_{s_1}^{s_2}g_{st}^{seg2}(\theta_{4-7})H_{s_2}^{s_3}g_{st}^{seg3}(\theta_{8-11})
\end{equation}
}

$g_{st}^{robot}(\theta)$ will yield the tool-frame position and orientation in $SE(3)$ format expressed in the robot coordinate frame. Robot-specific dimensional parameters and more details regarding the calculation process can be found in the Supplementary Materials. In the actual implementation, we use Wolfram Mathematica to pre-compute one analytic expression of the products of exponentials for all three segments. It is a function of segment dimension parameters and segment joint variables.

\subsection{Inverse Kinematics of the UMArm}

The inverse kinematics of UMArm is achieved through using the iterative inverse Jacobian method \cite{murray2017mathematical}. Each revolute joint on the arm has a \revcomment{1.9}{$\pm 0.3~rad~(17.2\degree)$} range of motion. In this work, the inverse kinematics problem will be solved for the position of the end-effector, while we assume that the orientation will be achieved with extra DoFs on the end-effector. ``Vanilla'' Jacobian inverse kinematics cannot limit the joint angle or smartly distribute motion among the 12 DoF. As a result, the IK solver will output solutions that are not feasible for our physical system. Modifications need to be implemented to the Jacobian IK solver. UMArm has many redundancies granted by the large number of independently actuated muscles. This makes the null space of the manipulator Jacobian non-empty, which effectively means that the robot can change its posture in specific ways while still maintaining the end-effector position. We can use this feature to our advantage to redistribute motion reasonably among the joints. We have implemented a method that combines the gradient projection method \cite{liegeois1977automatic} and the weighted least norm method \cite{chan1995weighted}. These modifications will encourage each joint to stay within the range of motion limit. The detailed Jacobian derivation and the gradient projection method can be found in the Supplementary Materials.

\begin{figure}
    \centering
    \includegraphics[width=\linewidth]{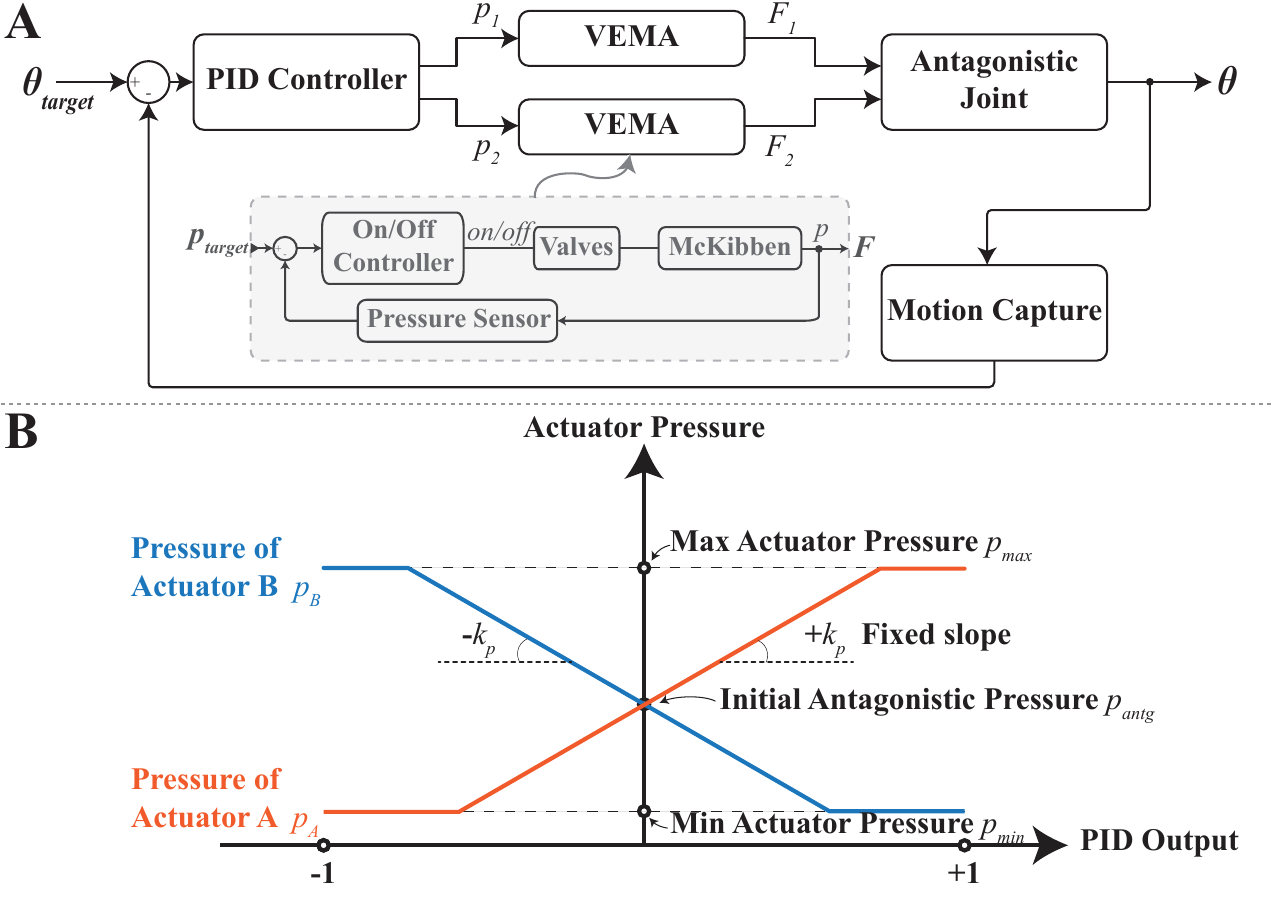}
    \caption{\textbf{A:} Closed loop controller diagram of the joint PID controller and the actuator internal pressure on-off controller.
    \revcomment{0}{\textbf{B:} Mapping from the PID controller output to the pressure inside the two antagonistic pair actuators.}}
    \label{fig:pid}
\end{figure}

\subsection{Joint Angle Controller}

\revcomment{0}{Each revolute joint is designed to be controlled by two VEMAs as described in Fig. \ref{fig:structure}C. We use a standard PID controller for each revolute joint to control the joint angle in a closed-loop manner. As shown in the control diagram in Fig. \ref{fig:pid}A, the PID controller acquires the actual joint angle feedback from the motion capture system. According to the error input and the target joint angle setpoint, the PID controller outputs a value between -1 and 1. Since two actuators control one joint, we need to convert the PID output into actuator pressures.
}

\revcomment{0}{Figure \ref{fig:pid}B shows the method of mapping PID output onto actuator pressures. The two curves represent the pressure inside the two antagonistic pair actuators. At zero PID output, both actuators start with the initial antagonistic pressure $p_{antg}$. As the PID output change, depending on the direction, one actuator pressure will increase toward the maximum actuator pressure $p_{max}$, while the other actuator will decrease toward the minimun actuator pressure $p_{min}$. The change rate $k_p$ is set to a constant value.}

\revcomment{1}{
\begin{equation}
    \begin{cases}
    p\ts{A} = p\ts{antg} + k\ts{p} \cdot PID\ts{out} \\
    p\ts{B} = p\ts{antg} - k\ts{p} \cdot PID\ts{out} \\
    p\ts{A} = min(p\ts{max}, p\ts{A}) \\
    p\ts{B} = max(p\ts{min}, p\ts{B})
    \end{cases}
\end{equation}
}

\revcomment{0}{The stiffness of each joint can be tuned by simultaneously (antagonistically) increasing/decreasing the pressure inside the two actuators. To do that, the arm user can set the initial antagonistic stiffness for each joint $p_{antg}$. The mapping shown above is a simple and balanced method for converting PID output to actuator pressures. It is also possible to generate different mapping schemes that fit the need for different applications.}

\begin{figure}
    \centering
    \includegraphics[width=\linewidth]{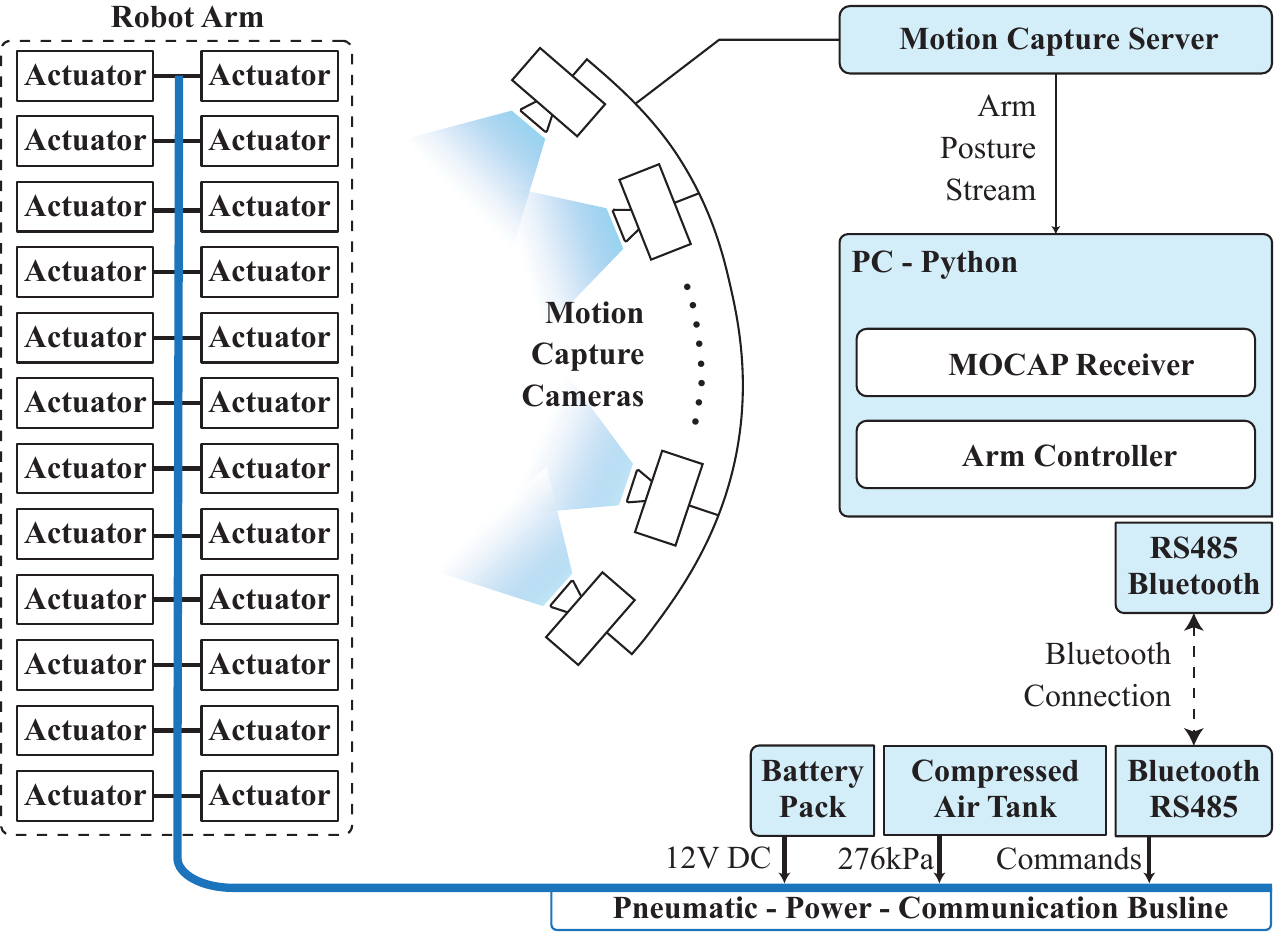}
    \caption{Schematic of the experimental setup. The actuators (from our previous work \cite{zuo2024embedded}) are connected to a bus line that receives voltage from a battery pack, pneumatic pressure from a compressed air tank, and command signals wirelessly via Bluetooth. The arm posture is captured by a 12-camera motion capture system. A laptop is running a Python application that processes the arm posture stream and formulates arm control commands.}
    \label{fig:schematic}
\end{figure}

\section{Characterization of the Arm}

\subsection{Experiment Setup}

In the characterization experiments, the arm's base is mounted to the ceiling rail, with the arm naturally hanging down as the rest configuration. The arm and its peripheral system are connected as shown in Figure \ref{fig:schematic}. The arm operates using a laboratory compressed air outlet and a 12V DC source. The two-line RS485 serial communication interface connects to an RS485-to-USB adapter, then plugs into the PC. A 12-camera motion capture system \revcomment{3.5}{(OptiTrack Flex 13)} is used to track the deformation of the arm. Motion capture markers are installed on cantilever arms on the printed link frame shown in Fig. \ref{fig:structure}C. Four markers on one link frame define one rigid body. In each rigid body, the four markers form a quadrilateral, and the intersection point of its diagonals aligns with the center of the U-joint. The 3-segment arm has six rigid bodies in total, and the rigid bodies' positions and orientations can be acquired from the motion capture system. By using the rigid body position and orientation, we can compute each joint angle defined in Figure \ref{fig:structure}A. A multi-process Python application runs on the laptop PC. One process receives the data stream from the motion capture server, and for each frame, calculates the joint angle of the robot. Another process runs the 12-joint PID controller and the serial interface that sends target pressures to the actuators. A third process is dedicated to the computationally heavy kinematics functions, and the last process runs an interface that receives user commands and visualizes the robot. In the characterization section, a 0.1m rod with 4 motion capture markers (shown in Fig.\ref{fig:structure}A) is installed as the end-effector.

\subsection{Step Response of the Arm}

\begin{figure}
    \centering
    \includegraphics[width=\linewidth]{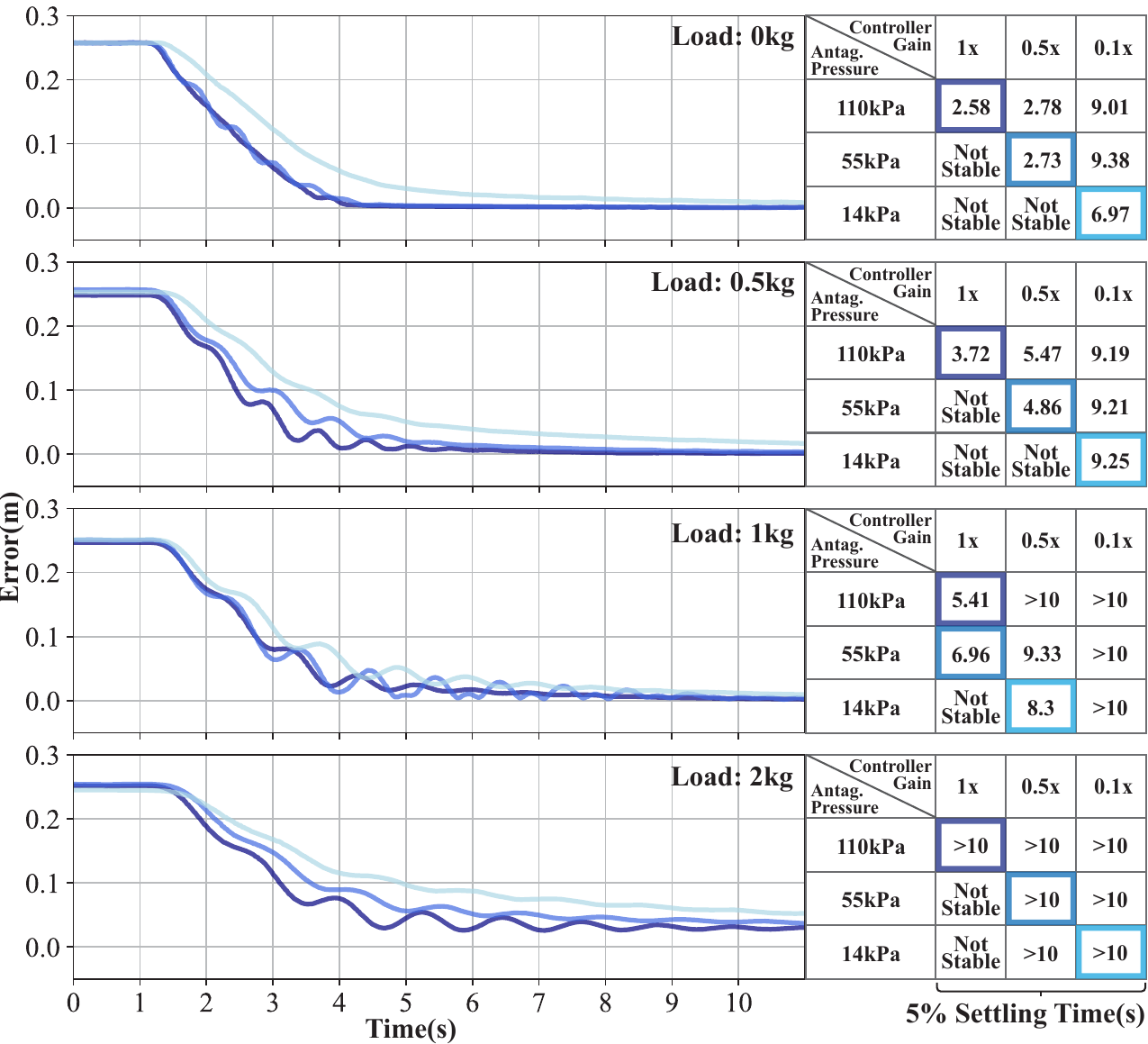}
    \caption{\revcomment{1.2, 1.10}{Position control step response of the UMArm. At $t=0$, the arm starts from its resting position and is commanded to reach a target position using the inverse kinematics controller. Four load conditions are tested. For each load, the effects of antagonistic pressure and controller gain are evaluated. The 5\% settling times are summarized in the accompanying table, and for each antagonistic pressure setting, the best-performing trial is plotted. Line colors correspond to those in the table.}}
    \label{fig:ikine_step}
\end{figure}

\revcomment{1.2}{The performance of the joint angle controller and the inverse kinematics position controller is evaluated through a step response experiment. At $t=0$~s, the inverse kinematics controller is activated (which also utilizes the joint angle controller), and the target position is set to the position at the initial configuration with all joints at zero. Subsequently, a new target position located 0.25~m from the initial position is commanded, and the UMArm attempts to reach this position with its end-effector.}

\revcomment{1.2}{Figure~\ref{fig:ikine_step} presents the results of the step response experiment under four loading conditions: 0~kg, 0.5~kg, 1~kg, and 2~kg. To evaluate the effect of antagonistic stiffening, three levels of structural stiffness were tested for each load by varying the antagonistic pressure ($p\ts{antg}$) to 110~kPa, 55~kPa, and 14~kPa. In each case, the chosen $p_{antg}$ was applied uniformly across all 12 pairs of antagonistic actuators.
}

\revcomment{1.2}{The tables accompanying each plot summarize the experimental settings. The first column lists the antagonistic pressure ($p\ts{antg}$) applied to each joint, which determines the structural stiffness. Because changes in stiffness also affect system dynamics, the global gain of the joint angle controller was adjusted accordingly. For each configuration, controller gains of 1$\times$, 0.5$\times$, and 0.1$\times$ were tested in the step response experiments. The gains are directly applied after the PID output.}

\revcomment{1.2}{We computed the 5\% settling time for each trial, and the results are displayed in the tables shown on the right side of each plot. For every load condition and antagonistic pressure, the plot displays the best-performing trial, defined as the one with the shortest settling time. The data curves represent the development of the root-mean-square error (RMSE) between the target position and the actual position over time. At zero load, $p_{antg}=110kPa$, and 1$\times$ controller gain, UMArm can achieve a steady-state positional error of 0.85~mm.}

Qualitatively speaking, a joint with higher antagonistic stiffness will act like a spring with a large spring constant. Upon disturbances, its response will have a lower magnitude. Also, a McKibben actuator at higher pressure can dissipate energy faster due to the larger internal friction between the fibers. \revcomment{1.11}{Therefore, the joint can structurally lower the oscillation magnitude and increase the damping factor. As a result, the arm can quickly stabilize when a new target position is given.
}

\subsection{\revcomment{0}{Positional Control Precision}}

\begin{figure}
    \centering
    \includegraphics[width=\linewidth]{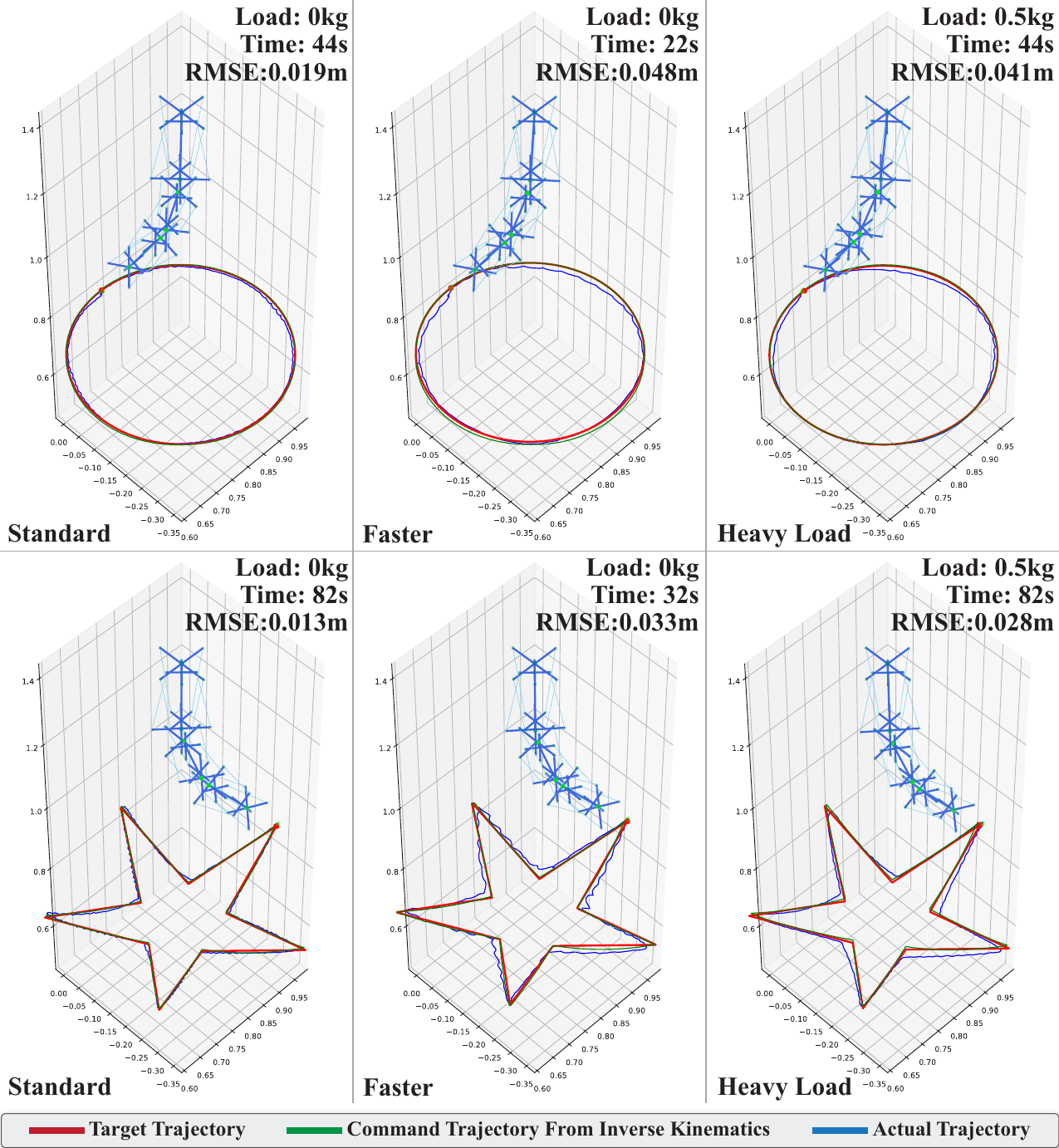}
    \caption{\revcomment{1.4, 1.5, 4.7}{The UMArm is tasked with drawing circular and star-shaped patterns in the spatial coordinate system. The target trajectory is shown in red, with a point moving along it at constant speed. The commanded trajectory from the inverse kinematics controller is shown in green, and the actual end-effector trajectory recorded by the motion capture system is shown in blue.}}
    \label{fig:pattern_drawing}
\end{figure}

\begin{figure}
    \centering
    \includegraphics[width=\linewidth]{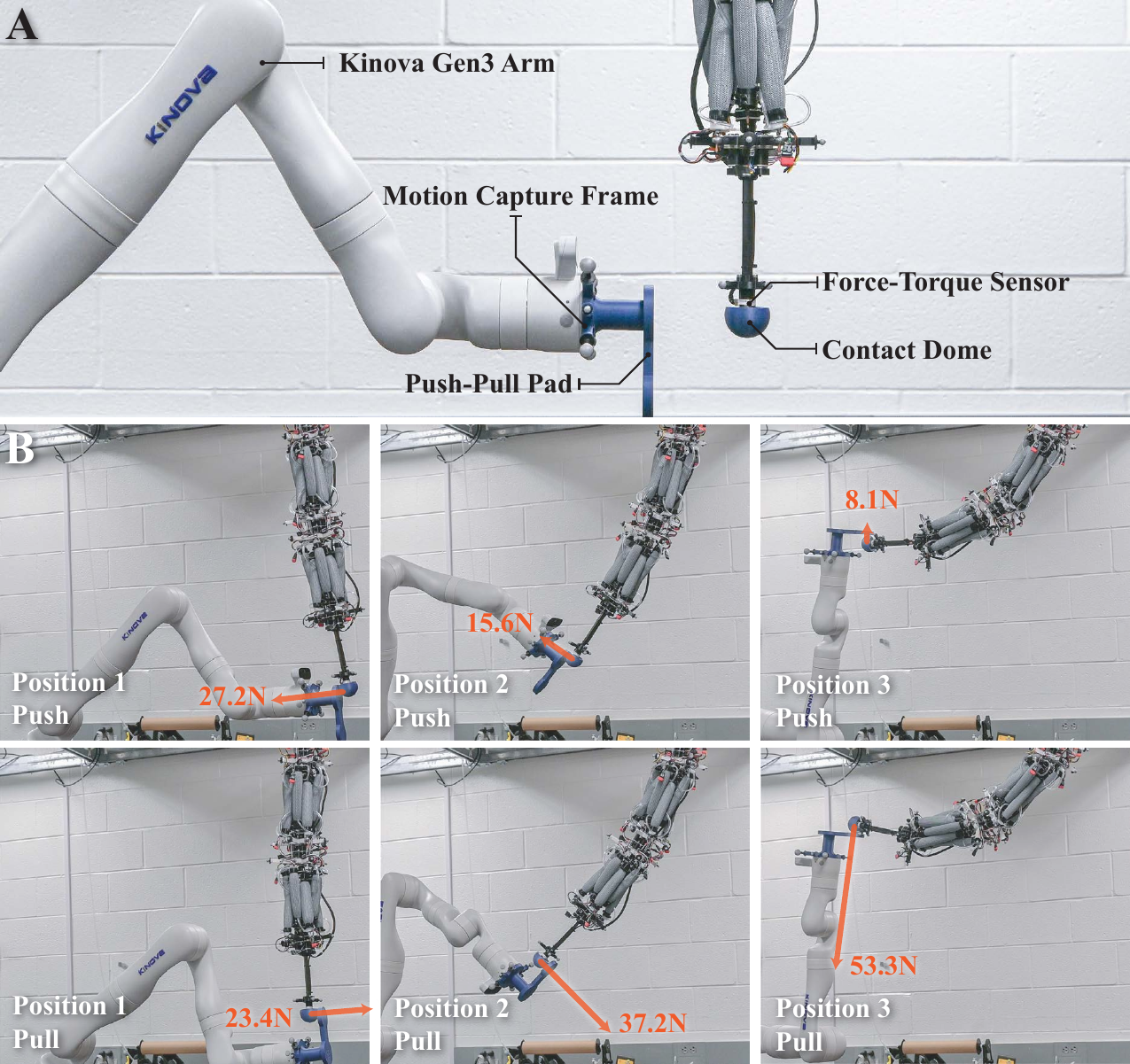}
    \caption{\revcomment{1.5, 2.3}{(A) Experimental setup for testing UMArm’s force generation at different configurations. (B) The UMArm is first commanded to hold its end-effector at a fixed position. A Kinova Arm then applies push and pull motions to displace the end-effector from this position. In response, the UMArm exerts force on the push–pull pad as it attempts to return to the commanded position. The maximum generated force is measured by the force–torque sensor and plotted in each figure. This procedure is repeated for three different arm configurations.}}
    \label{fig:force_transmission}
\end{figure}

\revcomment{4.4}{We first evaluated the UMArm’s inverse kinematics controller in simulation, without connecting to the physical hardware. A total of 1000 points were randomly sampled within the UMArm workspace, and the UMArm was tasked with traversing these waypoints using its end-effector. A computation time of 0.5~s was allowed between consecutive waypoints. The resulting position errors showed strong accuracy: 50\% of the points were within 0.1~mm, 91\% within 1~mm, 99\% within 2~mm, and all points within 3~mm.}

\revcomment{4.7}{Next, we conducted a trajectory tracking experiment with the UMArm hardware. As shown in Figure~\ref{fig:pattern_drawing}, the arm was tasked with drawing (circle and star) in the spatial coordinates. The desired trajectory was first generated, and then a target point was moved along the pattern at constant speed. The inverse kinematics controller then guided the UMArm’s end-effector to follow this moving target. The figure plots the reference pattern, the commanded trajectory from the controller, and the actual trajectory recorded by the motion capture system.  
}

\revcomment{4.7}{
We tested three conditions for each pattern: (i) no load at standard speed, (ii) no load at a higher speed, and (iii) 500~g load at standard speed. The tracking error is defined as the Euclidean distance between the target position and the actual position measured by motion capture. Because the joint controller is PID-based, some error is unavoidable, primarily due to the lag of the actual position behind the moving target. Nonetheless, the tracking remains smooth and stable, and the overall patterns drawn are very close to the intended trajectories.}

\subsection{\revcomment{0}{Force Generation Capability}}

\revcomment{2.3}{The force generation capability of the UMArm is characterized experimentally. Figure~\ref{fig:force_transmission}A shows the setup. A force–torque sensor (ATI Nano-17) is mounted on the UMArm, with a spherical contact dome attached to the sensor. A Kinova Gen3 arm is then used to push and pull the contact dome. Three end-effector positions are tested, and at each position, the UMArm’s ability to apply force in both pushing and pulling directions is evaluated. 
}

\revcomment{2.3}{
Figure~\ref{fig:force_transmission}B illustrates the experimental procedure. The UMArm first uses its inverse kinematics controller to move the end-effector to one of three predefined positions. The Kinova Arm then displaces the end-effector by 50~mm, either pushing or pulling it away from the target position. Owing to its controller design, the UMArm attempts to restore the end-effector position, thereby applying a reaction force normal to the push–pull pad of the Kinova Arm. This force is measured by the force–torque sensor.  
}

\revcomment{2.3}{
At position~1, the arm generates similar force outputs in both pushing and pulling directions. At position~2, it produces a greater force in pulling and a reduced force in pushing, while at position~3 this asymmetry is even more pronounced. The difference arises from actuator length asymmetry: at position~1, all actuators are approximately equal in length, whereas at positions~2 and~3 the actuators responsible for pushing are shortened while those for pulling are stretched. Since McKibben actuators generate stronger forces when stretched at the same pneumatic pressure, larger pulling forces than pushing forces are observed at positions~2 and~3.
}

\subsection{\revcomment{0}{Untethered Operation}}

The arm can operate untethered using a portable pressure source. We demonstrate this functionality using an off-the-shelf compressed air tank that holds 1.11L (68ci) at 31000kPa (4500 psi) (Fig \ref{fig:centipede}A). It was originally built for airsoft players, so it is designed to be portable and lightweight. We characterize the untethered performance by letting the robot traverse the eight way-points with two 2-second intervals.\revcomment{0}{The way-point arrangement is shown in Fig. 2 in the Supplementary Material.} The experiment results show that the compressed air tank we use could supply the robot to traverse 700 waypoints, or run continuously for 11min 45s before it starts to lose tracking accuracy due to the depletion of the tank.

\section{UMArm Demonstrations}

\subsection{Variable Structural Stiffness}

\begin{figure}
    \centering
    \includegraphics[width=\linewidth]{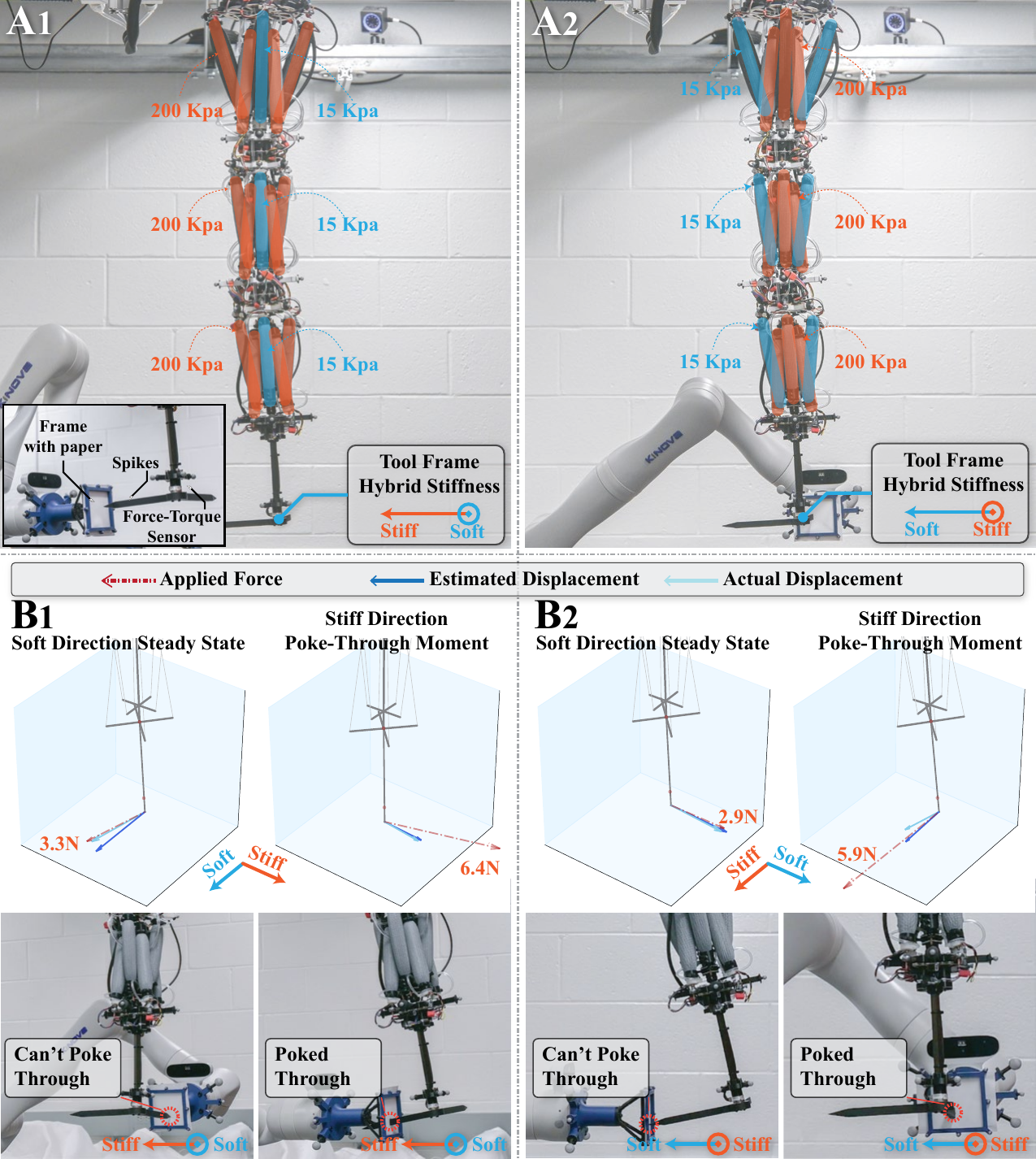}
    \caption{\revcomment{1.5, 2.4, 2.6, 2.7, 4.6}{Testing UMArm's variable structural stiffness ability. (A1, A2). The McKibben actuators are selectively pressurized for different tool-frame stiffness profiles. A Kinova Gen3 Arm is equipped with a frame with printer paper installed. A 2-direction spike is installed on the force-torque sensor on the end-effector of the UMArm. (B1, B2). The Kinova Arm pushes the paper frame against the 2-direction spike. The spike can easily penetrate the paper along the stiff direction, but cannot penetrate the paper along the soft direction. The force for each case is recorded by the force torque sensor. The skeleton model plots also show the estimated vs actual displacement from the applied external force.}}
    \label{fig:variable_stiffness}
\end{figure}

\begin{figure}
    \centering
    \includegraphics[width=\linewidth]{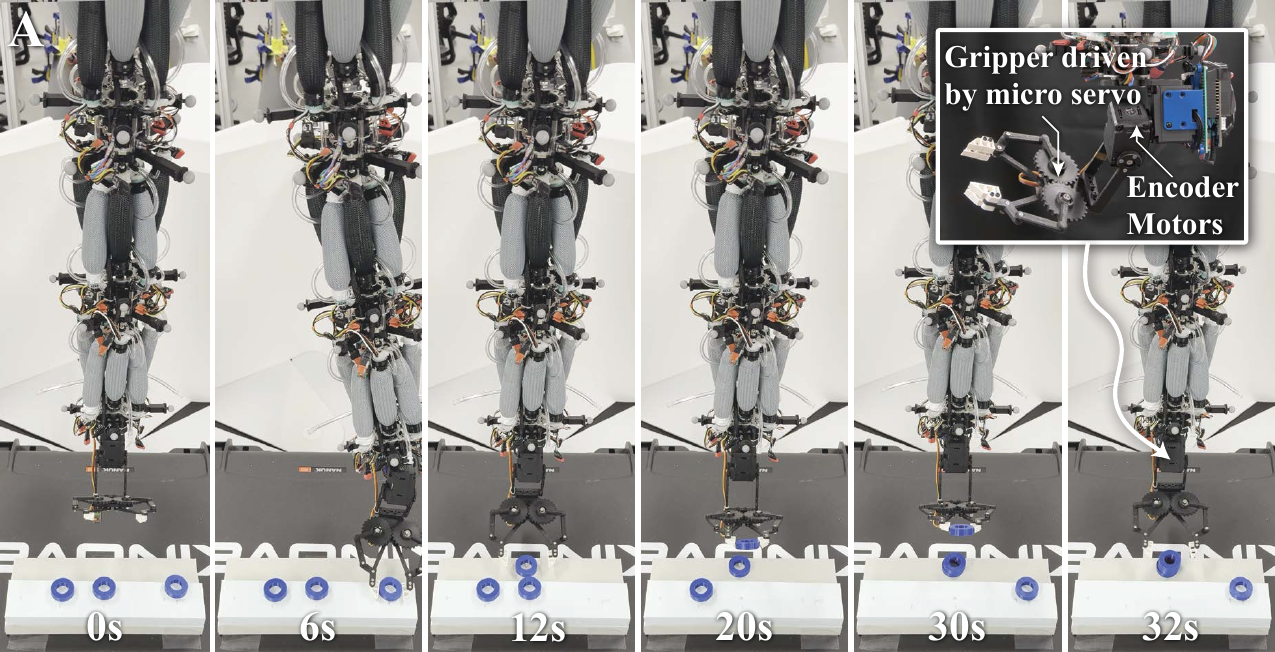}
    \caption{\revcomment{2.9}{A 2-DoF end-effector module consists of 2 encoder motors and a gripper is installed on the arm. Using the arm's current configuration, the controller will calculate encoder motor angles so that the motors can maintain the spatial orientation of the gripper. The gripper is driven by a mini servo. \textbf{B:} With the installed gripper end-effector, the arm can stack the three objects together. In the fifth and sixth sub-figures, an extra object is placed ahead of time by a person for the next round of stacking.}}
    \label{fig:stack}
\end{figure}

\begin{figure}
    \centering
    \includegraphics[width=\linewidth]{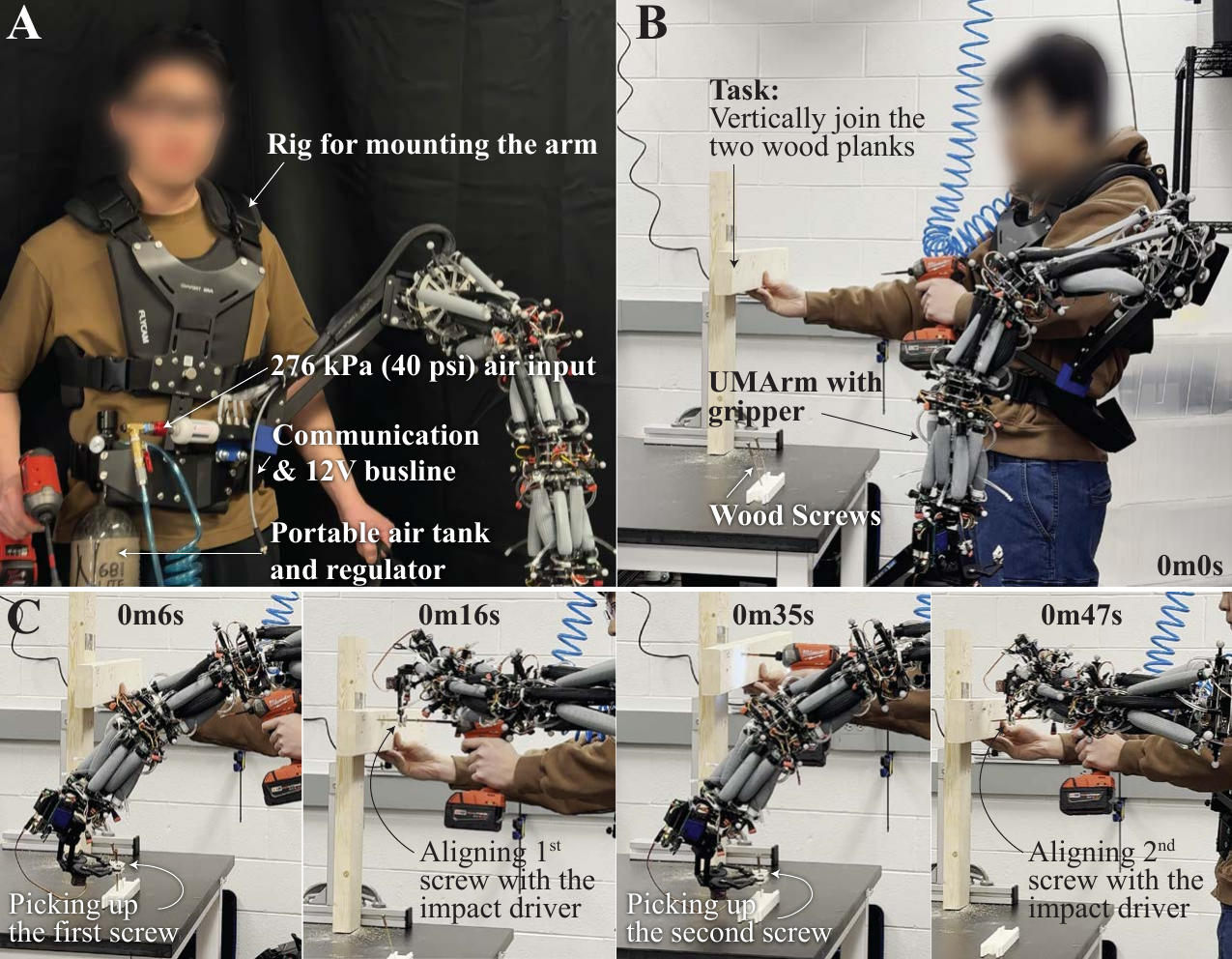}
    \caption{\textbf{A:} Front and side views of the wearable assistive arm setup. The UMArm is attached to a rig worn by the human operator. \textbf{B:} The worker is tasked with perpendicularly joining two heavy wooden planks and requires an extra arm to hold the wood screws. \textbf{C:} The assistive arm grasps the screw from the table, aligns it with the worker's impact driver, and repeats the process for the second screw.}
    \label{fig:wearable}
\end{figure}

\revcomment{2.4, 2.7, 4.6}{
Soft robotic arms are structurally compliant, which is crucial for safe interaction with humans. However, for a robotic arm to be precise and fast, it needs to have control over its structural stiffness. Otherwise, we might see oscillatory behavior or buckling under heavy load. UMArm can control the structural compliance on all 12 joints on the arm. This has granted the UMArm great flexibility to have safety and performance simultaneously. Figure \ref{fig:variable_stiffness} demonstrates UMArm's directional compliance control capability.}

\revcomment{2.4, 2.7, 4.6}{
We 3D printed a two-direction spike and mounted it perpendicularly on the end-effector. By selectively stiffening certain joints and relaxing others, the UMArm was configured to be structurally stiff along one direction while remaining soft along the other direction. To observe this anisotropic stiffness, a Kinova Arm held a sheet of printer paper and pressed it against the spike.
}

\revcomment{2.4, 2.7, 4.6}{
Figure~\ref{fig:variable_stiffness} A1/B1 show the UMArm under the first stiffness profile, where the end-effector is stiff in the left–right direction and soft in the front–back direction. A2/B2 illustrate the inverse profile. In both cases, the printer paper was not penetrated when pressed against the soft direction, but was penetrated when pressed against the stiff direction.
}

\revcomment{2.4, 4.6}{
To further analyze UMArm’s variable stiffness capability, we would like to develop a modeling approach to quantitatively estimate directional compliance as a function of the arm’s configuration and actuator pressures. The structure of UMArm is fully compatible with the stiffness estimation method proposed by Bruder et al.~\cite{bruder2023increasing}. We adapted the principles of their approach and implemented a three-dimensional version tailored to UMArm.
}

\revcomment{2.4, 4.6}{
Using the configuration vector \textbf{q} and the actuator pressure vector \textbf{p}, we compute the task-space compliance matrix $C\ts{x}(\textbf{q}, \textbf{p}) \in \mathbb{R}^{3\times3}$. The derivation of $C\ts{x}$ is provided in the Supplementary Materials. This matrix can be used to predict the virtual displacement $\delta\textbf{x} \in \mathbb{R}^{3}$ of the end-effector under an external virtual force $\delta F\ts{ext} \in \mathbb{R}^{3}$:
}

\revcomment{1}{
\begin{equation}
    \delta\textbf{x} = C\tb{x}(\textbf{q,p})\delta F\ts{ext}
\end{equation}
}

\revcomment{2.4, 4.6}{
During the paper-penetration test, external force data were collected from the force–torque sensor, while displacement data were obtained from the motion capture system. These measurements were used to validate the variable stiffness model. In Figure~\ref{fig:variable_stiffness}~B1/B2, the skeleton model of UMArm prior to each paper press is shown in gray. For the soft direction, force and displacement are measured after the system stabilizes; for the stiff direction, they are measured just before the spikes penetrate the paper. Externally applied forces are plotted as orange dashed arrows. Although the skeleton model is shown in the pre-displacement state, the plotted force vectors were recorded after displacement. The post-displacement force vector is transformed in the spatial frame to preserve its relative alignment with the robot. The actual displacement vector is shown as a light blue arrow, starting from the pre-displacement end-effector position and ending at the post-displacement position. The task-space compliance matrix $C\ts{x}$ is computed using the pre-displacement configuration \textbf{q}; when multiplied by the plotted force vector, it yields the estimated displacement, shown as a medium blue arrow.}

\revcomment{2.4, 4.6}{
Although the estimated displacement is valid only locally, the results still reveal a clear trend: along the compliant direction, small forces produce relatively large displacements, whereas along the stiff direction, large forces result in only small displacements. These results confirm that the modeled structural compliance matrix provides a useful quantitative estimate of the UMArm’s end-effector stiffness profile.
}

\subsection{Precision Handling Arm}

Having many independently controlled actuators also grants the UMArm the ability to perform delicate tasks. It can move very small distances with great precision. We showcase that ability through an object-stacking experiment. \revcomment{0}{The top-right corner of Figure \ref{fig:stack} shows the end-effector we have developed for this task.} This end-effector consists of two encoder motors, one gripper driven by a servo, and the controller circuits. The two encoder motors can maintain the gripper's orientation. Figure \ref{fig:stack} shows the object stacking experiment. There are three objects on the upper level of a two-level platform. The UMArm picks up and places the first object onto the lower platform, stacks the second object on top of the first object, and stacks the third object on top of both objects. In the supplementary video, the robot can keep repeating this task nonstop with great consistency.

\subsection{Wearable Assistive Arm}

UMArm is very suitable to be a wearable assistive arm because of its lightweight design, untethered operation ability, large payload capacity, and dexterity. \revcomment{2.5}{Many previous works have explored the use of soft robots for wearable assistive arms \cite{Nguyen2019Fabric, Nguyen2019Soft, Hussain2016SixthFinger}. However, few designs aim to achieve portability, precision, payload capacity, dexterity, and a wide range of motion simultaneously, which are the goals addressed by UMArm.} Fig. \ref{fig:wearable}A shows a worker wearing the UMArm as a worksite assistive arm. A rig built for a camera stabilizer is repurposed to attach the arm to the worker. The task for the worker is to perpendicularly join two wooden planks using two wood screws. As the worker needs to use one hand to hold the plank and another hand to hold the impact driver, thus an extra hand is needed to hold the wood screw. Figure \ref{fig:wearable} B\&C demonstrates how the UMArm could assist in such a situation. The arm can use the gripper to pick up the first screw from the screw holder, and align the screw at where the worker wants it. As the worker starts to drill in the screw, the arm can automatically pick up the second screw, and align it again. With the help of UMArm, the worker is able to complete the task within one minute.

\subsection{Inchworm Robot}

The regulator-embedded design, untethered operation ability, and large number of independently controlled actuators have made the UMArm extremely versatile. We want to demonstrate the versatility by showing that the UMArm can not only serve as a robotic arm, but also operate as a standalone robot that can untetheredly locomote and interact. \revcomment{2.5}{
While many soft robots have demonstrated impressive locomotion capabilities through diverse mechanisms \cite{ahmad2018kirigami, qi2023bioinspired, qin2018design}, UMArm demonstrates that untethered locomotion and interaction can be achieved simultaneously using the same set of actuators.}

\begin{figure}
    \centering
    \includegraphics[width=\linewidth]{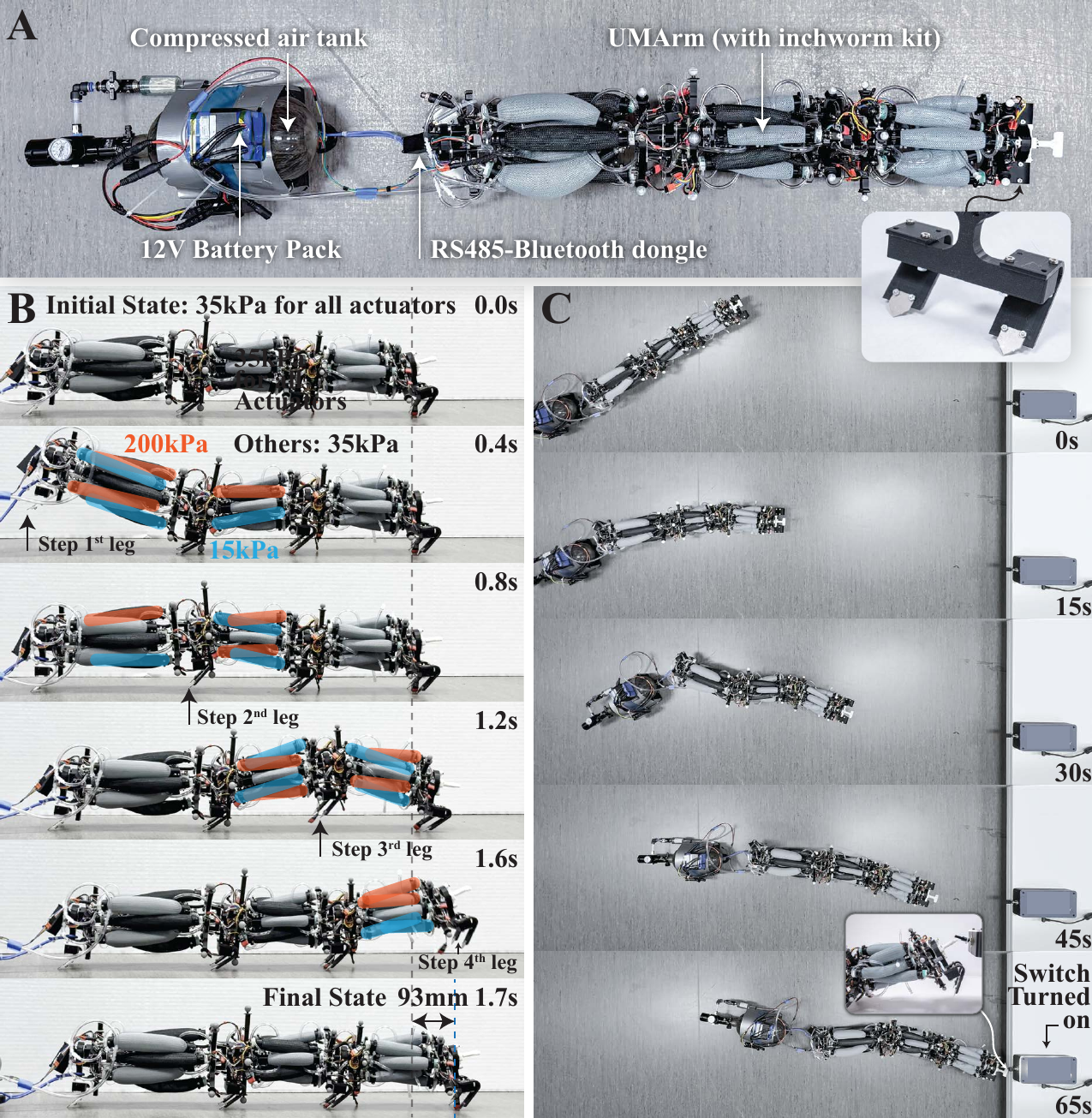}
    \caption{\textbf{A:} Feet are added in between each link to transform the UMArm into a bio-mimetic inchworm robot. The compressed air tank and a battery pack are loaded onto a trailer and towed along with the robot. \textbf{B:} Shows one inchworming cycle. The robot sequentially raises and steps its feet to walk forward. \textbf{C:} The inchworm robot is tasked to turn on a switch. It crawls to the light switch and raises its first segment to toggle the on switch. \textit{Note: Closed-up shots taken separately from the demo.}}
    \label{fig:centipede}
\end{figure}

With some slight modifications, we turned the UMArm into an untethered inchworm robot. Figure \ref{fig:centipede}A shows the modified UMArm. Four legs are installed on the UMArm. The legs are metal spikes that have a 45\degree{} angle with the ground. It has less traction sliding forward and larger traction sliding backward. The compressed air tank and 12V battery pack are loaded on a wheeled trailer, which will be tagged along with the inchworm robot. Figure \ref{fig:centipede}B shows the gaits of the inchworm. It raises and steps forward the legs one at a time. When inchworming forward, it can complete one cycle in 1.7 seconds, during which it moves forward by 93 millimeters. A more comprehensive demonstration of its ability is shown in Figure \ref{fig:centipede}C. The inchworm is tasked with turning on the flip switch of a lightbox. It rotates and moves toward the switch. After it aligns its head with the switch, it needs to raise the first segment to flip the switch. However, if the other two segments are relaxed, as the first segment tries to exert force at the switch, the leg between the second and third segments will be lifted up, therefore, not enough force will be exerted onto the switch. Fortunately, the robot can employ the antagonistic stiffening feature to increase the stiffness of the body. By stiffening the body, the two segments at the back will stay straight and provide firm support for the raised first segment, which can now apply enough force to turn on the switch.

\section{Discussion}

\subsection{Discussion on Experimental Results}

\revcomment{1.3}{In the inverse kinematics control experiment, we evaluated the arm’s performance in moving from an initial position to a target setpoint. We observed a clear pattern: a heavier payload slows down convergence and introduces more instabilities, but increasing the structural stiffness improves settling time and enhances stability. In the force transmission experiment, the arm produced the expected results. The maximum force was comparable when the arm was straight. However, when the arm was bent (positions 2 and 3), it generated a greater pulling force than pushing force. This behavior arises from the mechanics of the McKibben actuator: when stretched, the actuator produces significantly higher pulling force. In bent configurations, more actuators operate at their optimal length for pulling, but at a less favorable length for pushing. The maximum measured force was 53.3~N, while the arm itself weighs only 1.15~kg.}

\revcomment{1.3}{In the pattern drawing experiment, we demonstrated that the robot can accurately trace prescribed paths both with and without a payload. The controller used is a PID controller, which responds to the instantaneous error between the current and target positions, rather than planning motion in advance as in model predictive control. As a result, some tracking error is inevitable. Compared with \cite{bruder2021koopman}, the UMArm achieved an average error of 19~mm when tracing a circular trajectory under no load, while the arm by Bruder et al. reported 26~mm error. Moreover, UMArm operated on a substantially larger scale (trajectory radius 250~mm vs. 100~mm), making its relative performance even stronger. The reported error represents the distance between the target and end-effector positions at the same timestamp; thus, it does not fully capture trajectory similarity, which can be better appreciated by directly inspecting the plotted trajectories. Larger errors occur at sharp corners because the target position advances at constant speed along the trajectory, while the robot’s end-effector inevitably lags. Since the trajectory is not adjusted to wait for the end-effector, the robot’s motion appears less precise at these turns.}

\revcomment{1.3}{In the variable stiffness experiment, Figure~B compares the estimated displacement with the actual displacement. While the results are qualitatively similar, it is worth pointing out that they should not coincide in general. The estimated displacement corresponds to a virtual displacement under the action of a virtual force and is valid only locally. By contrast, the actual displacement vector is computed from the initial and displaced end-effector positions, which is a non-local measurement from the motion capture system. Discrepancies are therefore expected. More accurate estimation of the true displacement may be achieved with improved modeling, which may be addressed in future work.}

\subsection{Limitations and Future Works}

\iffalse
\revcomment{1.3}{
\dan{This opening paragraph should be removed, all the information in it gets repeated immediately in subsequent paragraphs (I think it's just an old version that wasn't deleted).}
% We would also like to share the limitations we discovered in the development process of the UMArm. 
Despited its demonstrated performance, UMArm has some notable limitations. 
First, the workspace of the UMArm is quite limited. Although UMArm has 12 independently driven revolute joints, each joint has a limited range of motion ($\pm{0.3rad}$). The workspace of the UMArm looks like an eggshell with limited depth. To resolve this hardware limitation, a linear rail powered by a stepper motor can be added to the base of the UMArm, to allow actuation in the $+z$ axis. This simple modification can greatly enlarge the volume of the workspace, while not undermining the key contribution of the system, such as variable structural stiffness. Second, we recognize the limitations of the current PID controller. We anticipate that advanced control techniques tailored to this hardware system could better exploit the UMArm’s capabilities and provide faster and more efficient responses with appropriately tuned antagonistic pressure.}
\fi

\revcomment{1.3}{Several limitations were identified during the development of UMArm. First, its workspace is relatively restricted compared to conventional rigid robot arms. Although UMArm has 12 independently actuated revolute joints, each has a limited range of motion ($\pm 0.3$~rad), resulting in a workspace resembling a shallow eggshell (workspace plot available in the Supplementary Materials). To address this limitation, a linear rail with a stepper motor could be integrated at the base to provide motion along the $+z$ axis. This simple addition would significantly enlarge the workspace without diminishing key features such as variable structural stiffness.}

\revcomment{1.3, 3.4}{Second, we recognize the limitations of the current PID controller. In this work, our primary goal was to demonstrate the hardware capabilities of UMArm under a simple controller. Advanced control strategies tailored to the UMArm hardware may better leverage its unique properties, achieving faster, more efficient responses by intelligently tuning the large amount of independently tunable actuator pressures.}

\revcomment{1.3, 3.5}{Third, the absence of internal joint angle sensing restricts the arm to environments equipped with external motion capture. Incorporating embedded joint angle feedback would allow operation beyond the motion capture area. Specifically, we can build our robot joints using Metal 3D printing, and Hall-effect angle encoders will be integrated onto the joints. The encoder can detect the rotation angle of the magnetic shaft, thereby acquiring the joint angles.}

\revcomment{1.3, 3.3}{Fourth, while UMArm currently relies on a rigid spine with revolute joints to ensure stability and precise force transmission, this architecture necessarily limits the range of continuous deformations compared to fully soft or continuum arms. It could be helpful to investigate hybrid strategies that increase structural compliance while retaining robustness and localized controllability. Examples include integrating variable-stiffness metamaterials into the links \cite{yitengma} or adopting compliant flexure-based joints instead of rigid pin joints. Importantly, we do not view uncontrolled compliance as a desirable goal, since it can lead to buckling, oscillation, and reduced payload capacity. Rather, the focus is on introducing compliance in a structured and tunable manner, so that the arm can selectively soften when adaptability is needed and stiffen when precision and force transmission are required.}

\revcomment{1.3, 3.4}{In future work, we aim to address these limitations and extend UMArm’s control capabilities. We plan to work on the control aspect of the UMArm, which is particularly challenging due to the high degrees of freedom and the nonlinear behavior of soft actuators. Nonetheless, UMArm shows great potential: with a well-developed dynamical model and advanced motion planning, it could exploit its variable stiffness and compliance to achieve highly efficient, dynamic motions that are difficult to realize with conventional rigid robotic arms.}

\section{Conclusion}

This work introduces the UMArm, a pneumatically actuated rigid-soft hybrid arm that overcomes key limitations in soft robotic arms by integrating 24 independently controlled valve-embedded McKibben actuators. Our design achieves high precision \revcomment{0}{(end-effector steady-state RMSE 0.85 mm)}, strong force output (\revcomment{0}{53.3N while pulling at fully bend configuration}), and rapid actuation while remaining lightweight (1.15\,kg). Unlike traditional pneumatic arms requiring extensive external hardware, UMArm operates untethered using a compact, onboard pneumatic system, significantly enhancing portability. The modular architecture enables scalability and reconfiguration. The UMArm also demonstrates versatility. It can transform into an untethered inchworm robot that can move and interact. By leveraging antagonistic stiffening, the arm achieves directional compliance control. It can be tuned soft in one direction for safe human interaction, and rigid in the other direction for better force transmission. These advancements position UMArm as a robust solution for precise, wearable, and multimodal robotic applications. In the future, we would like to explore various control strategies that exploit the hardware advantages of the UMArm, striving for the goal of deploying high-performance, structurally compliant arms into the unstructured world.

\section{Acknowledgment}

This research is funded by the University of Michigan, Mechanical Engineering Department.
%% Optional (comment out if not in use)
% 

%% REFERENCES
\bibliographystyle{IEEEtran}
\bibliography{references}

@inproceedings{zuo2024embedded,
  title={Embedded Valves for Distributed Control of Soft Pneumatic Actuators},
  author={Zuo, Runze and Mehta, Mayank and Han, Dong Heon and Bruder, Daniel},
  booktitle={2024 IEEE/RSJ International Conference on Intelligent Robots and Systems (IROS)},
  pages={8286--8292},
  year={2024},
  organization={IEEE}
}

@article{bruder2021chain,
  title={The Chain-Link Actuator: Exploiting the Bending Stiffness of McKibben Artificial Muscles to Achieve Larger Contraction Ratios},
  author={Bruder, Daniel and Wood, Robert J},
  journal={IEEE Robotics and Automation Letters},
  volume={7},
  number={1},
  pages={542--548},
  year={2021},
  publisher={IEEE}
}

@article{bruder2021koopman,
  title={Koopman-based control of a soft continuum manipulator under variable loading conditions},
  author={Bruder, Daniel and Fu, Xun and Gillespie, R Brent and Remy, C David and Vasudevan, Ram},
  journal={IEEE Robotics and Automation Letters},
  volume={6},
  number={4},
  pages={6852--6859},
  year={2021},
  publisher={IEEE}
}

@article{bruder2018force,
  title={Force generation by parallel combinations of fiber-reinforced fluid-driven actuators},
  author={Bruder, Daniel and Sedal, Audrey and Vasudevan, Ram and Remy, C David},
  journal={IEEE Robotics and Automation Letters},
  volume={3},
  number={4},
  pages={3999--4006},
  year={2018},
  publisher={IEEE}
}

@article{zhang2020modular,
  title={Modular soft robotics: Modular units, connection mechanisms, and applications},
  author={Zhang, Chao and Zhu, Pingan and Lin, Yangqiao and Jiao, Zhongdong and Zou, Jun},
  journal={Advanced Intelligent Systems},
  volume={2},
  number={6},
  pages={1900166},
  year={2020},
  publisher={Wiley Online Library}
}

@inproceedings{almurib2012review,
  title={A review of application industrial robotic design},
  author={Almurib, Haider AF and Al-Qrimli, Haidar Fadhil and Kumar, Nandha},
  booktitle={2011 Ninth International Conference on ICT and Knowledge Engineering},
  pages={105--112},
  year={2012},
  organization={IEEE}
}

@article{brogaardh2007present,
  title={Present and future robot control development—An industrial perspective},
  author={Brog{\aa}rdh, Torgny},
  journal={Annual Reviews in Control},
  volume={31},
  number={1},
  pages={69--79},
  year={2007},
  publisher={Elsevier}
}

@article{zheng2020robust,
  title={Robust control of a silicone soft robot using neural networks},
  author={Zheng, Gang and Zhou, Yuan and Ju, Mingda},
  journal={ISA transactions},
  volume={100},
  pages={38--45},
  year={2020},
  publisher={Elsevier}
}

@article{rus2015design,
  title={Design, fabrication and control of soft robots},
  author={Rus, Daniela and Tolley, Michael T},
  journal={Nature},
  volume={521},
  number={7553},
  pages={467--475},
  year={2015},
  publisher={Nature Publishing Group UK London}
}

@article{wang2022control,
  title={Control strategies for soft robot systems},
  author={Wang, Jue and Chortos, Alex},
  journal={Advanced Intelligent Systems},
  volume={4},
  number={5},
  pages={2100165},
  year={2022},
  publisher={Wiley Online Library}
}

@article{kemp2007challenges,
  title={Challenges for robot manipulation in human environments [grand challenges of robotics]},
  author={Kemp, Charles C and Edsinger, Aaron and Torres-Jara, Eduardo},
  journal={IEEE Robotics \& Automation Magazine},
  volume={14},
  number={1},
  pages={20--29},
  year={2007},
  publisher={IEEE}
}

@article{le2020challenges,
  title={Challenges and conceptual framework to develop heavy-load manipulators for smart factories},
  author={Le, Chi Hieu and Le, Dang Thang and Arey, Daniel and Gheorghe, Popan and Chu, Anh My and Duong, Xuan Bien and Nguyen, Trung Thanh and Truong, Trong Toai and Prakash, Chander and Zhao, Shi-Tian and others},
  journal={International Journal of Mechatronics and Applied Mechanics},
  volume={8},
  number={2},
  pages={209--216},
  year={2020},
  publisher={CEFIN Publishing House}
}

@inproceedings{yang2019collaborative,
  title={Collaborative mobile industrial manipulator: a review of system architecture and applications},
  author={Yang, Manman and Yang, Erfu and Zante, Remi Christophe and Post, Mark and Liu, Xuefeng},
  booktitle={2019 25th international conference on automation and computing (ICAC)},
  pages={1--6},
  year={2019},
  organization={IEEE}
}

@article{li2024disturbance,
  title={Disturbance-adaptive tapered soft manipulator with precise motion controller for enhanced task performance},
  author={Li, Xianglong and Xiong, Quan and Sui, Dongbao and Zhang, Qinghua and Li, Hongwu and Wang, Ziqi and Zheng, Tianjiao and Wang, Hesheng and Zhao, Jie and Zhu, Yanhe},
  journal={IEEE Transactions on Robotics},
  year={2024},
  publisher={IEEE}
}

@article{ansari2017towards,
  title={Towards the development of a soft manipulator as an assistive robot for personal care of elderly people},
  author={Ansari, Yasmin and Manti, Mariangela and Falotico, Egidio and Mollard, Yoan and Cianchetti, Matteo and Laschi, Cecilia},
  journal={International Journal of Advanced Robotic Systems},
  volume={14},
  number={2},
  pages={1729881416687132},
  year={2017},
  publisher={SAGE Publications Sage UK: London, England}
}

@article{firth2022anthropomorphic,
  title={Anthropomorphic soft robotic end-effector for use with collaborative robots in the construction industry},
  author={Firth, Charlotte and Dunn, Kate and Haeusler, M Hank and Sun, Yi},
  journal={Automation in Construction},
  volume={138},
  pages={104218},
  year={2022},
  publisher={Elsevier}
}

@article{galloway2016soft,
  title={Soft robotic grippers for biological sampling on deep reefs},
  author={Galloway, Kevin C and Becker, Kaitlyn P and Phillips, Brennan and Kirby, Jordan and Licht, Stephen and Tchernov, Dan and Wood, Robert J and Gruber, David F},
  journal={Soft robotics},
  volume={3},
  number={1},
  pages={23--33},
  year={2016},
  publisher={Mary Ann Liebert, Inc. 140 Huguenot Street, 3rd Floor New Rochelle, NY 10801 USA}
}

@article{trivedi2008soft,
  title={Soft robotics: Biological inspiration, state of the art, and future research},
  author={Trivedi, Deepak and Rahn, Christopher D and Kier, William M and Walker, Ian D},
  journal={Applied bionics and biomechanics},
  volume={5},
  number={3},
  pages={99--117},
  year={2008},
  publisher={Taylor \& Francis}
}

@article{trimmer2013soft,
  title={Soft robots},
  author={Trimmer, Barry},
  journal={Current Biology},
  volume={23},
  number={15},
  pages={R639--R641},
  year={2013},
  publisher={Elsevier}
}

@article{bruder2020data,
  title={Data-driven control of soft robots using Koopman operator theory},
  author={Bruder, Daniel and Fu, Xun and Gillespie, R Brent and Remy, C David and Vasudevan, Ram},
  journal={IEEE Transactions on Robotics},
  volume={37},
  number={3},
  pages={948--961},
  year={2020},
  publisher={IEEE}
}

@article{xavier2022soft,
  title={Soft pneumatic actuators: A review of design, fabrication, modeling, sensing, control and applications},
  author={Xavier, Matheus S and Tawk, Charbel D and Zolfagharian, Ali and Pinskier, Joshua and Howard, David and Young, Taylor and Lai, Jiewen and Harrison, Simon M and Yong, Yuen K and Bodaghi, Mahdi and others},
  journal={IEEE Access},
  volume={10},
  pages={59442--59485},
  year={2022},
  publisher={IEEE}
}

@article{whitesides2018soft,
  title={Soft robotics},
  author={Whitesides, George M},
  journal={Angewandte Chemie International Edition},
  volume={57},
  number={16},
  pages={4258--4273},
  year={2018},
  publisher={Wiley Online Library}
}

@article{seleem2023recent,
  title={Recent developments of actuation mechanisms for continuum robots: A review},
  author={Seleem, Ibrahim A and El-Hussieny, Haitham and Ishii, Hiroyuki},
  journal={International Journal of Control, Automation and Systems},
  volume={21},
  number={5},
  pages={1592--1609},
  year={2023},
  publisher={Springer}
}

@article{li2021position,
  title={Position control for soft actuators, next steps toward inherently safe interaction},
  author={Li, Dongshuo and Dornadula, Vaishnavi and Lin, Kengyu and Wehner, Michael},
  journal={Electronics},
  volume={10},
  number={9},
  pages={1116},
  year={2021},
  publisher={MDPI}
}

@article{bruder2023increasing,
  title={Increasing the payload capacity of soft robot arms by localized stiffening},
  author={Bruder, Daniel and Graule, Moritz A and Teeple, Clark B and Wood, Robert J},
  journal={Science Robotics},
  volume={8},
  number={81},
  pages={eadf9001},
  year={2023},
  publisher={American Association for the Advancement of Science}
}

@article{guan2023trimmed,
  title={Trimmed helicoids: an architectured soft structure yielding soft robots with high precision, large workspace, and compliant interactions},
  author={Guan, Qinghua and Stella, Francesco and Della Santina, Cosimo and Leng, Jinsong and Hughes, Josie},
  journal={npj Robotics},
  volume={1},
  number={1},
  pages={4},
  year={2023},
  publisher={Nature Publishing Group UK London}
}

@book{murray2017mathematical,
  title={A mathematical introduction to robotic manipulation},
  author={Murray, Richard M and Li, Zexiang and Sastry, S Shankar},
  year={2017},
  publisher={CRC press}
}

@article{mitchell2019easy,
  title={An easy-to-implement toolkit to create versatile and high-performance HASEL actuators for untethered soft robots},
  author={Mitchell, Shane K and Wang, Xingrui and Acome, Eric and Martin, Trent and Ly, Khoi and Kellaris, Nicholas and Venkata, Vidyacharan Gopaluni and Keplinger, Christoph},
  journal={Advanced Science},
  volume={6},
  number={14},
  pages={1900178},
  year={2019},
  publisher={Wiley Online Library}
}

@article{yang2018design,
  title={Design and implementation of a soft robotic arm driven by SMA coils},
  author={Yang, Hao and Xu, Min and Li, Weihua and Zhang, Shiwu},
  journal={IEEE Transactions on Industrial Electronics},
  volume={66},
  number={8},
  pages={6108--6116},
  year={2018},
  publisher={IEEE}
}

@inproceedings{feng2021design,
  title={Design of a Compliant 7-DoF Power Soft Robot driven by Hydraulic Artificial Muscles},
  author={Feng, Yunhao and Ide, Tohru and Nabae, Hiroyuki and Endo, Gen and Sakurai, Ryo and Ohno, Shingo and Suzumori, Koichi},
  booktitle={The 32nd 2021 International Symposium on Micro-NanoMechatronics and Human Science},
  pages={1--4},
  year={2021},
  organization={IEEE}
}

@article{ohta2018design,
  title={Design of a lightweight soft robotic arm using pneumatic artificial muscles and inflatable sleeves},
  author={Ohta, Preston and Valle, Luis and King, Jonathan and Low, Kevin and Yi, Jaehyun and Atkeson, Christopher G and Park, Yong-Lae},
  journal={Soft robotics},
  volume={5},
  number={2},
  pages={204--215},
  year={2018},
  publisher={Mary Ann Liebert, Inc. 140 Huguenot Street, 3rd Floor New Rochelle, NY 10801 USA}
}

@article{liu2023design,
  title={Design and experiment of a novel pneumatic soft arm based on a deployable origami exoskeleton},
  author={Liu, Yuwang and Shi, Wenping and Chen, Peng and Yu, Yi and Zhang, Dongyang and Wang, Dongqi},
  journal={Frontiers of mechanical engineering},
  volume={18},
  number={4},
  pages={54},
  year={2023},
  publisher={Springer}
}

@article{oh2024hybrid,
  title={Hybrid hard-soft robotic joint and robotic arm based on pneumatic origami chambers},
  author={Oh, Namsoo and Lee, Haneol and Shin, Jiseong and Choi, Youngjin and Cho, Kyu-Jin and Rodrigue, Hugo},
  journal={IEEE/ASME Transactions on Mechatronics},
  year={2024},
  publisher={IEEE}
}

@article{zhou2024pneumatic,
  title={A Pneumatic-Actuated Flexible Robotic Arm With Rigid Backbone for High-Precision and Safe Manipulation},
  author={Zhou, Xuanyi and Wu, Changqu and Cai, Shibo and Bao, Guanjun},
  journal={IEEE/ASME Transactions on Mechatronics},
  year={2024},
  publisher={IEEE}
}

@article{hongwei2022kinematic,
  title={Kinematic modeling and control of a novel pneumatic soft robotic arm},
  author={Hongwei, LI and Yan, XU and Huxiao, YANG and others},
  journal={Chinese Journal of Aeronautics},
  volume={35},
  number={7},
  pages={310--319},
  year={2022},
  publisher={Elsevier}
}

@article{alessi2023ablation,
  title={Ablation study of a dynamic model for a 3d-printed pneumatic soft robotic arm},
  author={Alessi, Carlo and Falotico, Egidio and Lucantonio, Alessandro},
  journal={IEEE Access},
  volume={11},
  pages={37840--37853},
  year={2023},
  publisher={IEEE}
}

@article{zhu2024design,
  title={Design and motion analysis of soft robotic arm with pneumatic-network structure},
  author={Zhu, Yinlong and Wang, Tian and Gong, Weizhuang and Feng, Kai and Wang, Xu and Xi, Shuang},
  journal={Smart Materials and Structures},
  volume={33},
  number={9},
  pages={095038},
  year={2024},
  publisher={IOP Publishing}
}

@article{jiang2020design,
  title={Design, control, and applications of a soft robotic arm},
  author={Jiang, Hao and Wang, Zhanchi and Jin, Yusong and Chen, Xiaotong and Li, Peijin and Gan, Yinghao and Lin, Sen and Chen, Xiaoping},
  journal={arXiv preprint arXiv:2007.04047},
  year={2020}
}

@inproceedings{garcia2024design,
  title={Design, manufacturing, and open-loop control of a soft pneumatic arm},
  author={Garc{\'\i}a-Samart{\'\i}n, Jorge Francisco and Rieker, Adri{\'a}n and Barrientos, Antonio},
  booktitle={Actuators},
  volume={13},
  number={1},
  pages={36},
  year={2024},
  organization={MDPI}
}

@article{sokolov2023design,
  title={Design, modelling, and control of continuum arms with pneumatic artificial muscles: A review},
  author={Sokolov, Oleksandr and Ho{\v{s}}ovsk{\`y}, Alexander and Trojanov{\'a}, Monika},
  journal={Machines},
  volume={11},
  number={10},
  pages={936},
  year={2023},
  publisher={MDPI}
}

@article{chan1995weighted,
  title={A weighted least-norm solution based scheme for avoiding joint limits for redundant joint manipulators},
  author={Chan, Tan Fung and Dubey, Rajiv V},
  journal={IEEE transactions on Robotics and Automation},
  volume={11},
  number={2},
  pages={286--292},
  year={1995},
  publisher={IEEE}
}

@article{liegeois1977automatic,
  title={Automatic supervisory control of the configuration and behavior of multibody mechanisms},
  author={Liegeois, Alain and others},
  journal={IEEE transactions on systems, man, and cybernetics},
  volume={7},
  number={12},
  pages={868--871},
  year={1977}
}

@INPROCEEDINGS{ian2006octarm,
  author={McMahan, W. and Chitrakaran, V. and Csencsits, M. and Dawson, D. and Walker, I.D. and Jones, B.A. and Pritts, M. and Dienno, D. and Grissom, M. and Rahn, C.D.},
  booktitle={Proceedings 2006 IEEE International Conference on Robotics and Automation, 2006. ICRA 2006.}, 
  title={Field trials and testing of the OctArm continuum manipulator}, 
  year={2006},
  volume={},
  number={},
  pages={2336-2341},
  doi={10.1109/ROBOT.2006.1642051}}

@article{
ahmad2018kirigami,
author = {Ahmad Rafsanjani  and Yuerou Zhang  and Bangyuan Liu  and Shmuel M. Rubinstein  and Katia Bertoldi },
title = {Kirigami skins make a simple soft actuator crawl},
journal = {Science Robotics},
volume = {3},
number = {15},
pages = {eaar7555},
year = {2018},
doi = {10.1126/scirobotics.aar7555},
URL = {https://www.science.org/doi/abs/10.1126/scirobotics.aar7555},
eprint = {https://www.science.org/doi/pdf/10.1126/scirobotics.aar7555}
}

@INPROCEEDINGS{qin2018design,
  author={Qin, Yun and Wan, Zhenyu and Sun, Yinan and Skorina, Erik H. and Luo, Ming and Onal, Cagdas D.},
  booktitle={2018 IEEE International Conference on Soft Robotics (RoboSoft)}, 
  title={Design, fabrication and experimental analysis of a 3-D soft robotic snake}, 
  year={2018},
  volume={},
  number={},
  pages={77-82},
  doi={10.1109/ROBOSOFT.2018.8404900}}

@article{qi2023bioinspired,
author = {Qi, Xinda and Gao, Tong and Tan, Xiaobo},
title = {Bioinspired 3D-Printed Snakeskins Enable Effective Serpentine Locomotion of a Soft Robotic Snake},
journal = {Soft Robotics},
volume = {10},
number = {3},
pages = {568-579},
year = {2023},
doi = {10.1089/soro.2022.0051},
note ={PMID: 36454198}
}

@article{Nguyen2019Soft,
author = {Nguyen, Pham Huy and Sparks, Curtis and Nuthi, Sai G. and Vale, Nicholas M. and Polygerinos, Panagiotis},
title = {Soft Poly-Limbs: Toward a New Paradigm of Mobile Manipulation for Daily Living Tasks},
journal = {Soft Robotics},
volume = {6},
number = {1},
pages = {38-53},
year = {2019},
doi = {10.1089/soro.2018.0065},
}

@ARTICLE{Hussain2016SixthFinger,
  author={Hussain, Irfan and Salvietti, Gionata and Spagnoletti, Giovanni and Prattichizzo, Domenico},
  journal={IEEE Robotics and Automation Letters}, 
  title={The Soft-SixthFinger: a Wearable EMG Controlled Robotic Extra-Finger for Grasp Compensation in Chronic Stroke Patients}, 
  year={2016},
  volume={1},
  number={2},
  pages={1000-1006},
  doi={10.1109/LRA.2016.2530793}}

@INPROCEEDINGS{Nguyen2019Fabric,
  author={Nguyen, Pham H. and Imran Mohd, I. B. and Sparks, Curtis and Arellano, Francisco L. and Zhang, Wenlong and Polygerinos, Panagiotis},
  booktitle={2019 International Conference on Robotics and Automation (ICRA)}, 
  title={Fabric Soft Poly-Limbs for Physical Assistance of Daily Living Tasks}, 
  year={2019},
  volume={},
  number={},
  pages={8429-8435},
  doi={10.1109/ICRA.2019.8794294}}

@ARTICLE{yitengma,
author={Ma, Yiteng and Bruder, Daniel},
journal={IEEE Robotics and Automation Letters (Accepted but not yet published)},
title={A Variable-Stiffness Robotic Link Based on Rotating-Rectangle Auxetic Structures for Safe Human-Robot Interaction},
year={2025}
}
\end{document}